\documentclass[preprint,12pt,authoryear]{elsarticle}

\usepackage{color,soul}
\usepackage{amsmath,amsfonts,amssymb}
\usepackage[ruled,vlined,linesnumbered,titlenumbered]{algorithm2e}
\usepackage{booktabs}
\usepackage{multirow}
\usepackage{graphicx}
\usepackage{caption}
\usepackage{times}
\usepackage{subcaption}
\usepackage{xr}
\usepackage{makecell}%To keep spacing of text in tables
\usepackage[margin=3.5cm]{geometry}
\usepackage{hyperref}
\usepackage{enumitem}
\usepackage{bm}
\usepackage{MnSymbol,bbding}
\usepackage{xparse,tikz}

\definecolor{mycolor}{RGB}{255,178,178}

\NewDocumentCommand{\circleddefaults}{m}{\tikz[baseline=(char.base)]{
		\node[shape=circle, fill=mycolor, draw=mycolor, text=black, inner sep=1pt] (char) {#1};}}

\def\Vhrulefill{\leavevmode\leaders\hrule height 0.7ex depth \dimexpr0.4pt-0.7ex\hfill\kern0pt}

\usepackage{amsmath}
\DeclareMathOperator*{\argmax}{arg\,max}

\newcolumntype{L}[1]{>{\raggedright\let\newline\\\arraybackslash\hspace{0pt}}m{#1}}
\newcolumntype{C}[1]{>{\centering\let\newline\\\arraybackslash\hspace{0pt}}m{#1}}
\newcolumntype{R}[1]{>{\raggedleft\let\newline\\\arraybackslash\hspace{0pt}}m{#1}}

\newcommand\algname{LUNAR}

\begin{document}

\begin{frontmatter}
	\title{\algname: Cellular Automata for Drifting Data Streams}
		
	\author[label1]{Jesus L. Lobo\corref{cor1}}
	\author[label1,label2,label3]{Javier Del Ser} 
	\author[label4]{Francisco Herrera} 
	
	\address[label1]{TECNALIA, Basque Research and Technology Alliance (BRTA), 48160 Derio-Bizkaia, Spain}
	\address[label2]{University of the Basque Country UPV/EHU, 48013 Bilbao, Spain}
	\address[label3]{Basque Center for Applied Mathematics (BCAM), 48009 Bilbao, Spain}
	\address[label4]{Andalusian Research Institute in Data Science and Computational Intelligence (DaSCI), University of Granada, 18071 Granada, Spain}
	\cortext[cor1]{Corresponding author: jesus.lopez@tecnalia.com (Jesus L. Lobo). TECNALIA, E-700, 48160 Derio (Bizkaia), Spain. Tl: +34 946 430 50. Fax: +34 901 760 009.}
	
	\begin{abstract}
	With the advent of huges volumes of data produced in the form of fast streams, real-time machine learning has become a challenge of relevance emerging in a plethora of real-world applications. Processing such fast streams often demands high memory and processing resources. In addition, they can be affected by non-stationary phenomena (concept drift), by which learning methods have to detect changes in the distribution of streaming data, and adapt to these evolving conditions. A lack of efficient and scalable solutions is particularly noted in real-time scenarios where computing resources are severely constrained, as it occurs in networks of small, numerous, interconnected processing units (such as the so-called Smart Dust, Utility Fog, or Swarm Robotics paradigms). In this work we propose \texttt{\algname}, a \textit{streamified} version of cellular automata devised to successfully meet the aforementioned requirements. It is able to act as a real incremental learner while adapting to drifting conditions. Extensive simulations with synthetic and real data will provide evidence of its competitive behavior in terms of classification performance when compared to long-established and successful online learning methods.
	\end{abstract}
	
	
\end{frontmatter}
 
\section{Introduction}\label{intro}

Real-Time Analytics (RTA), also referred to as stream learning, acquired special relevance years ago with the advent of the Big Data era \citep*{laney20013d,MOA-Book-2018}, becoming one of its most widely acknowledged challenges. Data streams are the basis of the real-time analytics, composed by sequences of items, each having a timestamp and thus a temporal order, and arriving one by one. Due to the incoming sheer volume of data, real-time algorithms cannot explicitly access all historical data because the storage capacity needed for this purpose becomes unmanageable. Indeed, data streams are fast and large (potentially, infinite), so information must be extracted from them in real-time. Under these circumstances, the consumption of limited resources (e.g. time and memory) often implies sacrificing performance for efficiency of the learning technique in use. Moreover, data streams are often produced by non-stationary phenomena, which imprint changes on the distribution of the data, leading to the emergence of the so-called \textit{concept drift}. Such a drift causes that predictive models trained over data flows become eventually obsolete, and do not adapt suitably to new distributions. Therefore, these predictive models need to be adapted to these changes as fast as possible while maintaining good performance scores \citep*{gama2014survey}. 

For all these reasons, the research community has devoted intense efforts towards the development of Online Learning Methods (OLMs) capable of efficiently undertaking predictive tasks over data streams under minimum time and memory requirements \citep*{widmer1996learning,ditzler2015learning,webb2016characterizing,lu2018learning,losing2018incremental,lobo2020spiking}. The need for overcoming these setbacks stems from many real applications, such as sensor data, telecommunications, social media, marketing, health care, epidemics, disasters, computer security, electricity demand prediction, among many others \citep*{vzliobaite2016overview}. The Internet of Things (IoT) paradigm deserves special attention at this point \citep*{manyika2015unlocking,de2016iot}, where a huge quantity of data is continuously generated in real-time by sensors and actuators connected by networks to computing systems. Many of these OLMs are based on traditional learning methods (i.e. Naive Bayes, Support Vector Machines, Decision Trees, Gaussian methods, or Neural Networks \citep*{MOA-Book-2018}), which have been \textit{streamified} to make them work incrementally and fast. By contrast, other models are already suitable for mining data streams by design \citep*{cervantes2018evaluating}. Unfortunately, most existing RtML models show a high complexity and dependence on the value of their parameters, thereby requiring a costly tuning process \citep*{losing2018incremental}. In addition, some of them are neither tractable nor interpretable, which are features lately targeted with particularly interest under the eXplainable Artificial Intelligence (XAI) paradigm \citep*{alej2019explainable}. Nowadays, learning algorithms featuring these characteristics are still under active search, calling for new approaches that blow a fresh breeze of novelty over the field \citep*{gama2014survey,khamassi2018discussion}. 

Cellular Automata (CA) become fashionable with the Conway's Game of Life in 1970, but scientifically relevant after the Stephen Wolfram's study in $2002$ \citep*{wolfram2002new}. Despite they are not widely used in data mining tasks, the Fawcett's work \citep*{fawcett2008data} showed how they can turn into a simple and low-bias data mining method, robust to noise, and with competitive classification performances in many cases. Until now, their appearance on the RTA has been timid, without providing evidences of their capacity for incremental learning and drift adaptation. 

This work enters this research avenue by proposing the use of Cellular Automata for RTA. Our approach, hereafter coined as \textit{celluLar aUtomata for driftiNg dAta stReams} (\texttt{\algname}), capitalizes on the acknowledged capacity of CA to model complex systems from simple structures. We show the intersection of CA and RTA in the presence of concept drift, showing that CA are promising incremental learners capable of adapting to evolving environments. Specifically, we provide a method to transform a traditional CA into its \textit{streamified} version (\texttt{sCA}) so as to learn incrementally from data streams. We show that \texttt{\algname} performs competitively with respect to other online learners on several real-world datasets. More precisely, we will provide informed answers to the following research questions:
\begin{itemize}[leftmargin=*]
	\item \texttt{RQ1}: Does \texttt{sCA} act as a real incremental learner?
	\item \texttt{RQ2}: Can \texttt{sCA} efficiently adapt to evolving conditions?
	\item \texttt{RQ3}: Does our \texttt{\algname} algorithm perform competitively with respect to other consolidated RTA approaches reported in the literature?
\end{itemize} 

The rest of the manuscript is organized as follows: first, Section \ref{rel_work} provides a general introduction to CA, placing an emphasis on their historical role in the pattern recognition field, and concretely their relevance for stream learning. Next, Section \ref{mets} delves into the methods and the \texttt{\algname} approach proposed in this work. Section \ref{exps} introduces the experimental setup, whereas Section \ref{reses} presents and discusses the obtained results from such experiments. Finally, Section \ref{concs} draws conclusions and future research lines related to this work.

\section{Related Work}\label{rel_work}

Before going into the technical details of the proposed approach, we herein provide a historical overview of CA (Subsection \ref{limelight} ), along with a perspective on how these models have been progressively adopted for pattern recognition (Subsection \ref{ca_pattern} ) and, more lately, stream learning (Subsection \ref{ca_stream} ).

\subsection{The Limelight Shone Down on Cellular Automata}\label{limelight}

The journey of CA was initiated by John von Neumann \citep*{neumann1966theory} and Ulam for the modeling of biological self-reproduction. They became really fashionable due to the popularity of Conway's Game of Life introduced by Gardner \citep*{games1970fantastic} in the field of artificial life \citep*{langton1986studying}. Arguably, the most scientifically significant and elaborated work on the study of CA arrived in $2002$ with the thoughtful studies of Stephen Wolfram \citep*{wolfram2002new}. In recent years, the notion of complex systems proved to be a very useful concept to define, describe, and study various natural phenomena observed in a vast number of scientific disciplines. Despite their simplicity, they are able to describe and reproduce many complex phenomena \citep*{wolfram1984cellular} that are closely related to processes such as self-organization and emergence, often observed within several scientific disciplines: biology, chemistry and physics, image processing and generation, cryptography, new computing hardware and algorithms designs (i.e., automata networks \citep*{goles2013neural} or deconvolution algorithms \citep*{zenil2019causal}), among many others \citep*{ganguly2003survey,bhattacharjee2016survey}. CA have also been satisfactorily implemented for heuristics \citep*{nebro2009mocell} and job scheduling \citep*{xhafa2008efficient}. Besides, the capability to perform universal computation \citep*{cook2004universality} has been one of the most celebrated features of CA that has garnered the attention of the research community: an arbitrary Turing machine can be simulated by a cellular automaton, so universal computation is possible \citep*{wolfram2002new}.

Considering their parallel nature (which allows special-purposed hardware to be implemented), CA are also called to be a breakthrough in paradigms such as Smart Dust \citep*{warneke2001smart,ilyas2018smart}, Utility Fog \citep*{hall1996utility,dastjerdi2016fog}, Microelectromechanical Systems (MEMS or ``motes") \citep*{judy2001microelectromechanical}, or Swarm Intelligence and Robotics \citep*{ramos2003swarms,del2019bio}, due to their capability to be computationally complete. Microscopic sensors are set to revolutionize a range of sectors, such as space missions \citep*{niccolai2019review}. The sensing, control, and learning algorithms that need to be embarked in such miniaturized devices can currently only be run on relatively heavy hardware, and need to be refined. However, nature has shown us that this is possible in the brains of insects with only a few hundred neurons. CA can also be decisive in other paradigms such as Nanotechnology, Ubiquitous Computing \citep*{lopez2017ubiquitous}, and Quantum Computation (introduced by Feynman \citep*{feynman1986quantum}), particularly the Quantum CA \citep*{lent1993quantum,watrous1995one,adamatzky2018cellular}. Due to the miniaturization hurdles of these devices, CA for stream learning may become of interest when there is no enough capacity to store huges volumes of data, and the computational capacity is very limited. After all, it is therefore not surprising that CA have received particular attention since the early days of CA investigation, and that CA for stream learning allows us to move in the correct direction in these scenarios, which are not far off the near future \citep*{jafferis2019untethered}.

\subsection{Foundations of Cellular Automata}\label{automaton}

CA are usually described as discrete dynamical systems which present a universal computability capacity \citep*{wolfram2018cellular}. Their beauty lies in its simple local interaction and computation of cells, which results in a huge complex behavior when these cells act together. 

\begin{figure}
	\begin{subfigure}{\textwidth}
		\centering
		% include first image
		\includegraphics[width=0.65\linewidth]{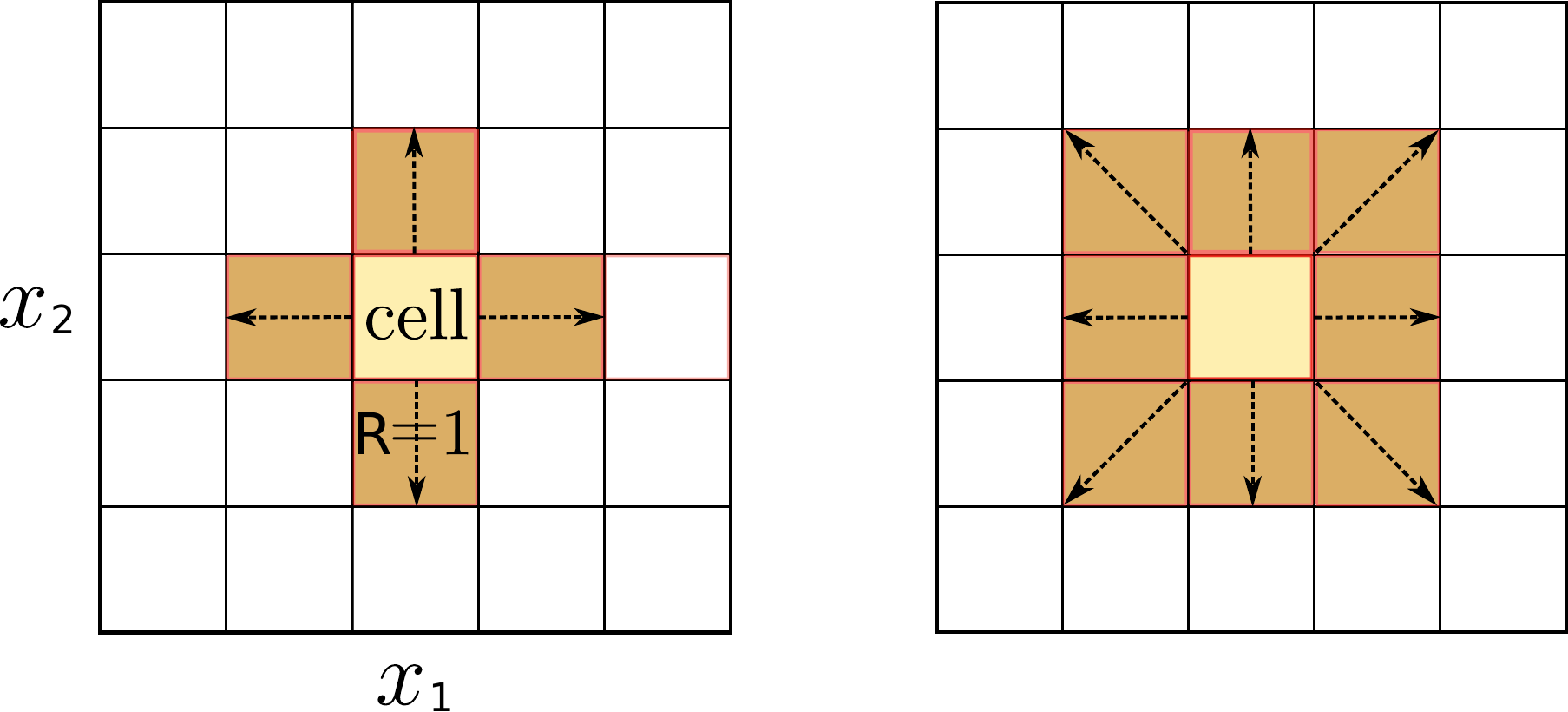}  
		\caption{}
		\label{fig_vn_vs_moore}
	\end{subfigure}
	\begin{subfigure}{.55\textwidth}
		\centering
		% include fourth image
		\includegraphics[width=\linewidth]{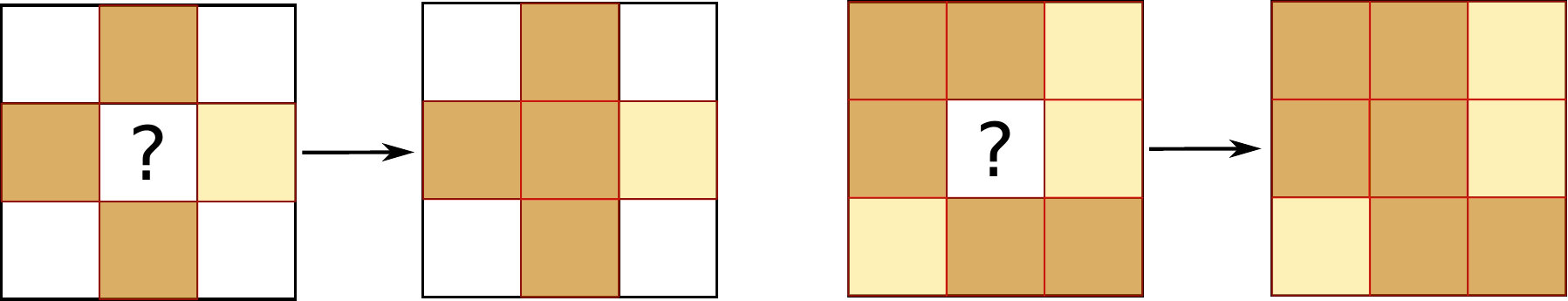}  
		\caption{}
		\label{fig_local_rule}
	\end{subfigure}
	\begin{subfigure}{.55\textwidth}
		\centering
		% include third image
		\includegraphics[width=.5\linewidth]{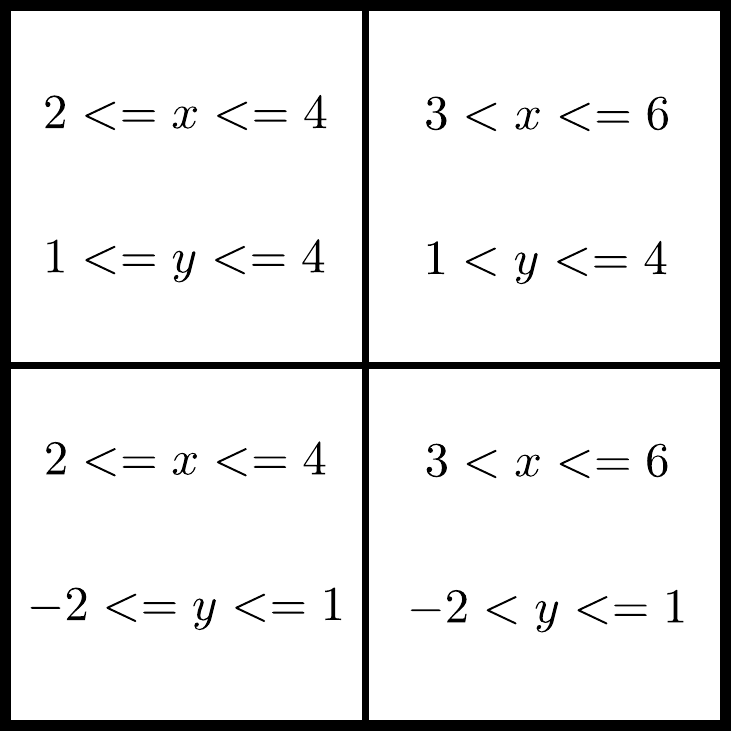}  
		\caption{}
		\label{fig_bins}
	\end{subfigure}
	\begin{subfigure}{\textwidth}
		\centering
		% include third image
		\includegraphics[width=\linewidth]{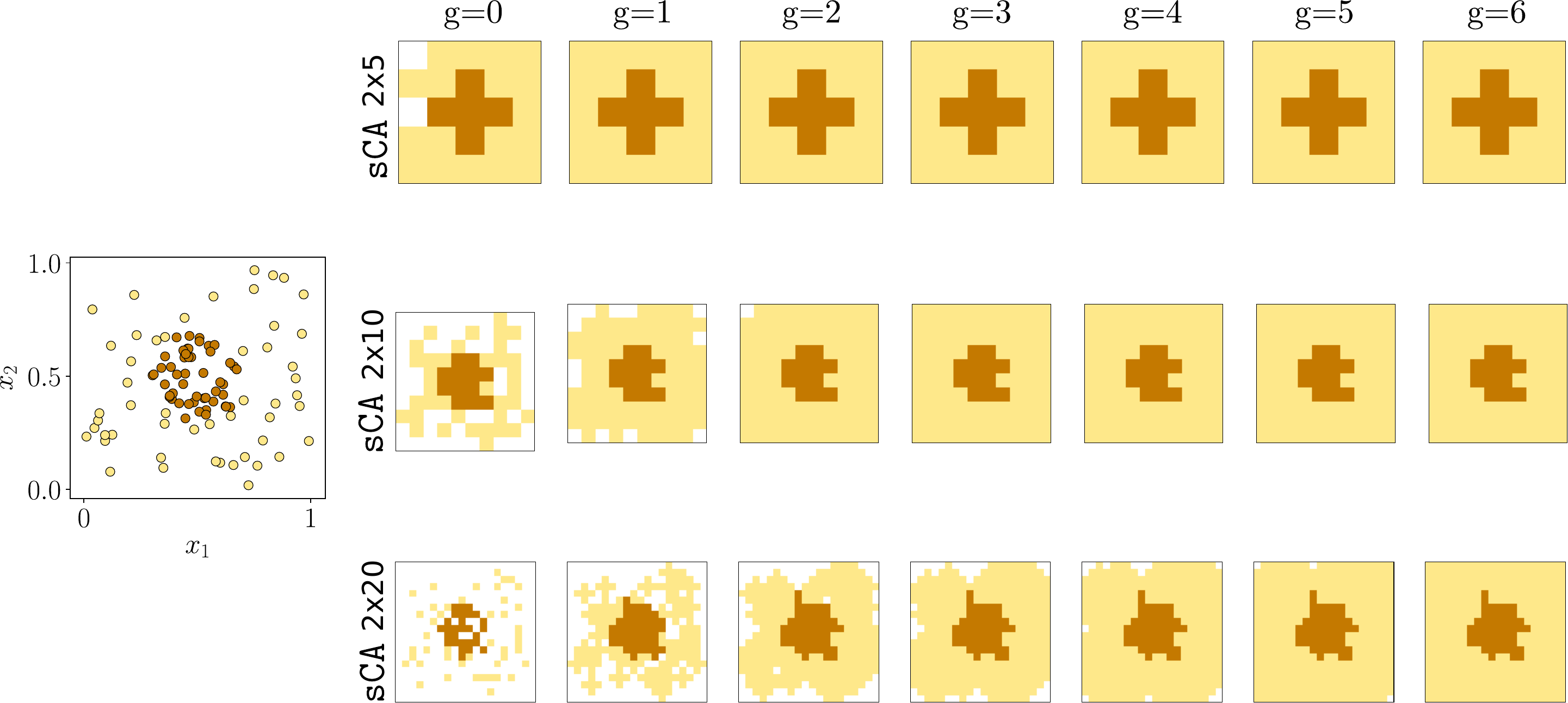}  
		\caption{}
		\label{fig_few_samples_yellow}
	\end{subfigure}	
	\caption{Visual representation of CA for pattern recognition: (a) the von Neumann's (left) and Moore's (right) neighborhoods with radius $R=1$; (b) the center cell examines its von Neumann's (left) and Moore's (right) neighborhoods and applies the local rule (majority vote) in a one-step update; (c) a two-dimensional dataset with instances $\mathbf{X}_t=(X_t^1,X_t^2)$ falling between $[2,6]$ (min/max $X_t^1$) and $[-2,-4]$ (min/max $X_t^2$), and a grid divided into $G=2$ bins; (d) von Neumann's automata ($R=1$, $d \times G$) ability to learn the data distribution from a few instances, and running up to $7$ iterations of the initialization step (generations).}
	\label{fig_fig_intro}
\end{figure}

We can find four mutually interdependent parts in CA: i) the lattice, ii) its states, iii) the neighborhood, and iv) the local rules. A \textit{lattice} is created by a grid of elements (cells), which can be composed in one, two or higher dimensional space, but typically composed of uniform squared cells in two dimensions. CA contain a finite set of discrete \textit{states}, whose number and range are dictated by the phenomenon under study. We find the simplest CA built using only one Boolean state in one dimension. The \textit{neighborhood}, which is used to evaluate a \textit{local rule}, is defined by a set of neighboring (adjacent) cells. A neighboring cell is any cell within a certain radius of the cell in question. Additionally, it is important to specify whether the radius $R$ applies only along the axes of the cell space (von Neumann neighborhood), or if it can be applied diagonally (Moore neighborhood). In two dimensions, the $R=1$ von Neumann neighborhood or the $R=1$ Moore neighborhood are often selected (see Figure \ref{fig_vn_vs_moore} ). Finally, a \textit{local rule} defines the evolution of each CA; it is usually realized by taking all states from all cells within the neighborhood, and by evaluating a set of logical or arithmetical operations written in the form of an algorithm (see Figure \ref{fig_local_rule} ). 
We can formally define a cellular automaton as follows, by adopting the notation used in \citep*{kari2005theory}: $A \doteq (d,\mathcal{S},f_{\boxplus},f_{\lcirclearrowdown})$, where $d$ represents the dimension, $\mathcal{S}$ a finite set of discrete states, $f_{\boxplus}(\cdot)$ is a function that given a cell's coordinates at its input, returns the neighbors of the cell to be used in the update rule, and $f_{\lcirclearrowdown}(\cdot)$ is a function that updates the state of the cell at hand as per the states of its neighboring cells. Therefore, in the case of a radius $R=1$ \emph{von Neumann} neighborhood defined over a $d=2$-dimensional lattice, the set of neighboring cells and state of cell with coordinates $\mathbf{c}=[i,j]$ is given by:
\begin{align}
&f_{\boxplus}([i,j])=\{[i,j+1],[i-1,j],[i,j-1],[i+1,j]\}, \\
&S(\mathbf{c})=S([i,j])=f_{\lcirclearrowdown}(S([i,j+1]),S([i-1,j]),S([i,j-1]),S([i+1,j])),
\end{align}
i.e., the states vector $S([i,j])$ of the $[i,j]$ cell within the lattice is updated according to the local rule $f_{\lcirclearrowdown}(\cdot)$ applied over its neighbors given by $f_{\boxplus}([i,j])$. In general, in a $d$-dimensional space, a cell's von Neumann neighborhood will contain $2d$ cells, and a Moore neighborhood will contain $3d-1$ cells. With this in mind, a cellular automaton should exhibit three properties to be treated as such: i) \textit{parallelism} or \textit{synchronicity} (all of the updates to the cells compounding the lattice are done at once); ii) \textit{locality} (when a cell $[i,j]$ is updated, its state $S[i,j]$) is based on the previous state of the cell and those of its nearest neighbors; and iii) \textit{homogeneity} or \textit{properties-uniformity} (the same update rule $f_{\lcirclearrowdown}(\cdot)$ is applied to each cell).
The use of CA for pattern recognition is not straightforward. Some modifications and considerations should be performed before using CA for pattern recognition, since we need to map a dataset to a cell space:
\begin{itemize}[leftmargin=*]
	\item \textit{Grid of cells} (lattice): despite being a natural fit to two-dimensional problems, CA must be extended to multiple dimensions to accommodate general pattern recognition tasks encompassing more dimensions. For $n$ features, an approach adopted in the related literature is to assign one grid dimension to each feature of the dataset. Once dimensions of the grid have been set, we need to partition each grid dimension by the feature's values, obtaining an equal number of cells per dimension. To do that, evenly spaced values based on the maximum and minimum feature values (with an additional margin (\textit{marg}) to ensure a minimal separation among feature's values) are used to create ``bins" for each dimension of the data (see Figure \ref{fig_bins} ), which ultimately yield the cells of the lattice.
	\item \textit{States}: a finite number of discrete states $|\mathcal{S}|$ are defined, corresponding to the number of classes considered in the task.
	\item \textit{Local rule}: in pattern recognition tasks the update rule $f(\cdot)$ can be set to very assorted forms, being the most straightforward a majority voting among the states (labels) of its neighbors, i.e. for $d=2$,
	\begin{equation}\label{eq_majvot}
	S([i,j])\hspace{-1mm}=\argmax_{s\in \mathcal{S}}\hspace{-3mm}\sum_{[k,l]\in f_{\boxplus}([i,j])} \hspace{-3mm}\mathbb{I}(S([k,l])=s),
	\end{equation}
	where $f_{\boxplus}([i,j])$ returns the coordinates of neighboring cells of $[i,j]$, and $\mathbb{I}(\cdot)$ is an auxiliary function taking value $1$ if its argument is true (and $0$ otherwise). Any other update rule can be defined to spread the state activation of each cell over its neighborhood (see Figure \ref{fig_local_rule} ).
	\item \textit{Neighborhood}: it is necessary to specify a neighborhood and its radius. Although there are more types of local rules, ``von Neumann'' or ``Moore'' neighborhoods are often used (Figure \ref{fig_vn_vs_moore} ).
	\item \textit{Initialization}: the grid is seeded depending on the feature values of the instances of the training dataset. Specifically, the state of each cell is assigned the label corresponding to the majority of training data instance with feature values falling within the range covered by the cell. As a result, cells will organize themselves into regions of similar labels (Figure \ref{fig_few_samples_yellow} ).
	\item \textit{Generations}: after the initialization step, some cells can remain unassigned. Then, it is necessary to run the CA (generations) until no cells are left empty, and then continuing until either no changes are made or a fixed threshold is exceeded. Along this process, each cell computes its new state by applying the update rule over the cells in its immediate neighborhood. Each cell follows the same update rule, and all cells are updated simultaneously and synchronously. A critical characteristic of CA is that the update rule examines only its neighboring cells, so the processing is entirely local. No global or macro grid characteristics are computed whatsoever (Figure \ref{fig_few_samples_yellow} ).
\end{itemize}

\subsection{Cellular Automata in Pattern Recognition}\label{ca_pattern}

Despite the early findings presented in \citep*{jen1986invariant,raghavan1993cellular,chaudhuri1997additive}, CA are not commonly used for pattern recognition. An exception is the work in \citep*{fawcett2008data}, where CA were used as a form of instance-based classifiers for pattern recognition. These kinds of classifiers are well known by the pattern recognition community in the form of instance-based learning and nearest neighbors classifiers \citep*{aha1991instance,duda2012pattern}. They represent regions of the instance space, so when a new instance needs to be classified, they select one or more close neighbors in these regions, and use their labels to assign a label to the new instance. Nevertheless, these are distinguished from CA in that they are not strictly local: there is no fixed neighborhood, hence an instance's nearest neighbor may change. In CA, by contrast, there is a fixed neighborhood, and the local interaction between cells influences the evolution and behavior of each cell. Fawcett, on the basis of Ultsch's work \citep*{ultsch2002data}, put CA in value for the pattern recognition community by introducing them as a low-bias data mining method, simple but powerful for attaining massively fine-grained parallelism, non-parametric, with an effective and competitive classification performance in many cases (similar to that produced by other complex data mining models), and robust to noise. Besides, they were found to perform well with relatively scarce data. All this prior evidence makes CA suited for pattern recognition tasks. 

Finally, it is worth highlighting the ability of CA to extract patterns from information and the possibility of generating tractable models, and being simple methods at the same time. This ability has encouraged the community to use them as a machine learning technique. Concretely, their traceability and reversibility  \citep*{kari2018reversible} are renowned drivers for their adoption in contexts where model explainability is sought \citep*{ribeiro2016should,gunning2017explainable,alej2019explainable}.

\subsection{Cellular Automata for Stream Learning}\label{ca_stream}

In a real-time data mining process (stream learning), data streams are read and processed once per arriving sample (instance). Algorithms learning from such streams (stream learners) must operate under a set of constrained conditions \citep*{domingos2003general}:
\begin{itemize}[leftmargin=*]
	\item Each instance can be processed only once.
	\item The processing time of each instance must be low.
	\item Memory must be low as well, which implies that only a few instances of the stream should be explicitly stored.
	\item The algorithm must be prepared for providing an answer (i.e. a prediction) at any time of the process.
	\item Data streams evolve along time, which is an aspect that should be mandatorily considered.
\end{itemize}

In mathematical terms, a stream learning process evolving over time can be formally defined as follows: given a time period $[0,t]$, we denote the historical set of instances as $\mathbf{D}_{0,t}={\mathbf{d}_{0},\ldots, \mathbf{d}_{t}}$, where $\mathbf{d}_{i}=(\mathbf{X}_{i},y_{i})$ is a data instance, $\mathbf{X}_{i}$ is the feature vector and $y_{i}$ its label. We assume that $\mathbf{D}_{0,t}$ follows a certain joint probability distribution $P_{t}(\mathbf{X},y)$. Such data streams are usually affected by non-stationary events (drifts) that eventually change their distribution (concept drift), making predictive models trained over these data obsolete. Bearing the previous notation in mind, concept drift at timestamp $t+1$ occurs if $P_{t}(\mathbf{X},y) \neq P_{t+1}(\mathbf{X},y)$, i.e. as a change of the joint probability distribution of $\mathbf{X}$ and $y$ at time $t$. 

Since in stream learning we cannot explicitly store all past data to detect or quantify this change, concept drift detection and adaptation are acknowledged challenges for real-time processing algorithms \citep*{lu2018learning}. Two strategies are usually followed to deal with concept drift:
\begin{itemize}[leftmargin=*]
	\item \emph{Passive}, by which the model is continuously updated every time new data instances are received, ensuring a sufficient level of diversity in its captured knowledge to accommodate changes in their distribution; and
	\item \emph{Active}, in which the model gets updated only when a drift is detected.
\end{itemize}

Both strategies can be successful in practice, however, the reason for choosing one strategy over the other is typically specific to the application. In general, a passive strategy has shown to be quite effective in prediction settings with gradual drifts and recurring concepts, while an active strategy works quite well in settings where the drift is abrupt. Besides, a passive strategy is generally better suited for batch learning, whereas an active strategy has been shown to work well in online settings \citep*{gama2004learning,bifet2007learning,alippi2013just}. In this work we have adopted an active strategy due to the fact that stream learning acts in an online manner.

Stream learning under non-stationary conditions is a field plenty of new challenges that require more attention due to its impact on the reliability of real-world applications. CA may provide a revolutionary view on RTA. Unfortunately, as shown later in Section \ref{automaton}, the original form of CA for pattern recognition does not allow for stream processing. A few timid attempts at incorporating CA concepts to RTA have been reported in the past. In \citep*{hashemi2007better,pourkashani2008cellular} a cellular automaton-based approach was used as a real-time instance selector for stream learning. The classification task is carried out in batch mode by other non-CA-based learning algorithms. This is an essential difference with respect to the \texttt{\algname} approach proposed in this work, as the CA approach in \citep*{hashemi2007better,pourkashani2008cellular} is not used for the learning task itself, but rather as a complement to the learning algorithm under use. Besides, they do not answer the relevant research questions posed in Section \ref{intro}, which should be the basis for the use of CA in stream learning. The developed \texttt{\algname} algorithm  is a \textit{streamified} learning version of CA. It transforms CA into real incremental learners that incorporate an embedded mechanism for drift adaptation. In what follows we address the posed research questions by providing rationale on the design of \texttt{\algname}, and discussing on a set of experiments specifically devoted to inform our answers.

\section{Proposed Approach: \algname}\label{mets}

As it has been previously mentioned, \texttt{\algname} relies on the usage of CA for RTA. In this section we first introduce the modifications to \emph{streamify} a CA pattern recognition approach (Subsection \ref{stream_CA} ). Secondly, \texttt{\algname} is deeply detailed, grounded on the previously introduced material (Subsection \ref{adapt_strat} ).

\subsection{Adapting Cellular Automata for Incremental Learning}\label{stream_CA}

The CA approach designed for pattern recognition can be adapted to cope with the constraints imposed by incremental learning. We present the details of this adaptation, which we coin as \emph{streamified} Cellular Automaton (\texttt{sCA}), in Algorithm \ref{alg_stream_automaton} (Streamified Cellular Automaton (\texttt{sCA})):
\begin{itemize}[leftmargin=*]
	\item First, the \texttt{sCA} is created after setting the value of its parameters as per the dataset at hand (lines $1$ to $5$). 
	\item Then, a set of $P$ preparatory data are used to initialize the grid off-line by assigning the states of the corresponding cells according to the instances' values (lines $6$ to $11$). 
	\item After this preliminary process, it is important to note that several preparatory data instances might collide into the same cell, thereby yielding several state occurrences for that specific cell. Since each cell must possess only one state for prediction purposes, lines $12$ to $16$ aim at removing this multiplicity of states by assigning each cell the state that occurs most among the preparatory data instances that fell within its boundaries. 
	\item Similarly, we must ensure that all cells have a state assigned, i.e. no empty grid cell status must be guaranteed. To this end, lines $17$ to $23$ apply the local rule $f_{\lcirclearrowdown}(\cdot)$ over the neighbors of every cell, repeating the process (generations) until every cell has been assigned a state. 
	\item This procedure can again render several different states for a given cell, so the previously explained majority state assignment procedure is again enforced over all cells of the grid (lines $24$ to $28$). 
\end{itemize}
Once this preparatory process is finished, the \texttt{sCA} is ready for the \textit{test-then-train} \citep*{gama2014survey} process with the rest of the instances (lines $29$ to $34$). In this process, the \texttt{sCA} first predicts the label of the arriving instance (testing phase), and updates the limits to reconfigure the bins (training phase). By this way, the \texttt{sCA} always represents the currently prevailing distribution of the streaming data at its input. After that, the \texttt{sCA} updates the current state of the cell enclosing the instance with the true label of the incoming instance. As a result, the cells of \texttt{sCA} always have an updated state according to the streaming data distribution.
\begin{algorithm}
	\DontPrintSemicolon
	\SetAlgoLined
	\SetKwInOut{Input}{Input}
	\SetKwInOut{Output}{Output}	
	\Input{Preparatory data instances $[(\textbf{X}_t,y_t)]_{t=0}^{t=P-1}$; training/testing data for the rest of the stream $[(\mathbf{X}_{t},y_t)]_{t=P}^{\infty}$; the grid size $G$ (bins per dimension); a local update rule \smash{$f_{\lcirclearrowdown}(\cdot)$}; a neighborhood function $f_{\boxplus}(\mathbf{c})$ for cell with coordinates $\mathbf{c}\in \mathcal{G}=\{1,\ldots,G\}^d$ as its argument; a radius $R$ for the neighborhood operator}	
	\Output{Trained streamified CA (\texttt{sCA}), producing predictions $\widehat{y}_t$ $\forall t\in[P,\infty)$}
	Let the number of dimensions $d$ of the grid be the number of features in $\mathbf{X}_t$\; 
	Let the number of cell states $|\mathcal{S}|$ be the number of classes (alphabet of $y_t$)\;	
	Set an empty vector of state hits per every cell: $\mathbf{h}_\mathbf{c}=[]$ $\forall \mathbf{c}\in\mathcal{G}$\;
	Initialize the limits of the grid: \smash{$[(lim_n^{low},lim_n^{high})]_{n=1}^d$}\;
	Create the \texttt{sCA} grid as per $G$, $n$ and \smash{$[(lim_n^{low},lim_n^{high})]_{n=1}^d$}\;
	\For(\tcp*[h]{Preparatory process}){$t=0$ to $P-1$}{
		Update limits as per $\mathbf{X}_{t}$, e.g., $lim_n^{low}=\min\{lim_n^{low},x_t^n\}$\;
		Reconfigure grid bins as per $G$ and the updated \smash{$[(lim_n^{low},lim_n^{high})]_{n=1}^d$}\;
		Select the cell $\mathbf{c}$ in the grid that encloses $\mathbf{X}_{t}$\;
		Append $y_t$ to the vector of state hits in the cell, e.g. $\mathbf{h}_{\mathbf{c}'}=[\mathbf{h}_{\mathbf{c}'},y_t]$\; 	
	}
	\For(\tcp*[h]{Guaranteeing one state per cell}){$\mathbf{c}\in \mathcal{G}$}{
		\If(\tcp*[h]{More than 1 state hit in cell}){$|\mathbf{h}_{\mathbf{c}}|>1$}{
			$S(\mathbf{c})=\argmax_{s\in\mathcal{S}}\smash{\sum_{i=1}^{|\mathbf{h}_{\mathbf{c}}|}} \mathbb{I}(h_{\mathbf{c}}^i=s)$\;
		}
	}	 
	\While(\tcp*[h]{Ensuring a hit in all cells}){$\exists \mathbf{c}\in\mathcal{G}:\:h_\mathbf{c}=[]$}{
		\For{$\mathbf{c}\in\mathcal{G}$}{
			Compute the neighboring cells of $\mathbf{c}$ as per $f_\boxplus(\mathbf{c})$ and $R$\;
			Compute state $s'$ by applying rule \smash{$f_{\lcirclearrowdown}(\cdot)$} over the neighboring cells\;
			Append $s'$ to \smash{$h_\mathbf{c}$}, i.e., $\mathbf{h}_{\mathbf{c}}=[\mathbf{h}_{\mathbf{c}},s']$\;
			
		}
	}
	\For(\tcp*[h]{Guaranteeing one state per cell}){$\mathbf{c}\in \mathcal{G}$}{
		\If(\tcp*[h]{More than 1 state hit in cell}){$|\mathbf{h}_{\mathbf{c}}|>1$}{
			$S(\mathbf{c})=\argmax_{s\in\mathcal{S}}\smash{\sum_{i=1}^{|\mathbf{h}_{\mathbf{c}}|}} \mathbb{I}(h_{\mathbf{c}}^i=s)$\;
		}
	}	
	\For(\tcp*[h]{Stream learning}){$t=P$ to $\infty$}{
		Predict $\widehat{y}_t$ as $S(\mathbf{c})$, with $\mathbf{c}$ denoting the coordinates of the cell enclosing $\mathbf{X}_t$\;
		Update limits as per $\mathbf{X}_{t}$, e.g., $lim_n^{low}=\min\{lim_n^{low},x_t^n\}$\;		
		Reconfigure grid bins as per $G$ and the updated \smash{$[(lim_n^{low},lim_n^{high})]_{n=1}^d$}\;
		Update cell state to the verified class of the test instance: $S(\mathbf{c})=y_t$\;
		}
	\caption{\textit{Streamified} Cellular Automaton (\texttt{sCA})}\label{alg_stream_automaton}
\end{algorithm}

The details of the proposed \texttt{\algname} algorithm underline the main differences between the CA version for pattern recognition and \texttt{sCA}. We highlight here the most relevant ones:
\begin{itemize}[leftmargin=*]
	\item \textit{Preparatory instances}: a small portion of the data stream (preparatory instances $[(\mathbf{X}_t,y_t)]_{t=0}^{P-1}$) is used to perform the \texttt{sCA} initialization. Then, the grid is seeded with these data values, and \texttt{sCA} progresses through several generations (and applying the local rule) until all cells are assigned a state. With the traditional CA for pattern recognition, the whole dataset is available from the beginning. Therefore, the CA initialization is not required. 
	\item \textit{An updated representation of the instance space}: since historical data is not available and data grow continuously, the \texttt{sCA} updates the cell bins of the grid every time a new data instance arrives. By doing this, we ensure that the range of values of the instance space is updated during the streaming process. With the traditional CA for pattern recognition, as the whole training dataset is static, the grid boundaries and cell bins are calculated at the beginning of the data mining process, and remain unaltered during the test phase.
	\item \textit{CA's cells continuously updated}: \texttt{sCA}'s cells are also updated when new data arrives by assigning the incoming label (state) to the corresponding cell. Instead of examining the cells in the immediate neighborhood of this corresponding cell during several generations (which would take time and computational resources), \texttt{sCA} just checks the previous state of the cell and the current state provided by the recent data instance. When both coincide, no change is required, but when they differ, the cell adopts the state of the incoming data instance. In this sense, \texttt{sCA} always represents the current data distribution. This is the way \texttt{sCA} works as an incremental learner. However, with the traditional CA for pattern recognition, once initialized, a local rule is applied over successive generations to set the state of a given cell depending on the states of its neighbors.  
\end{itemize}	

\subsection{\texttt{\algname}: a \texttt{sCA} with Drift Detection and Adaptation Abilities}\label{adapt_strat}

\texttt{\algname} blends together \texttt{sCA} and paired learning in order to cope with concept drift. In essence, a stable \texttt{sCA} is paired with a reactive one. The scheme inspires from the findings reported in \citep*{bach2008paired}: a stable learner is incrementally trained and predicts considering all of its experience during the streaming process, whereas a reactive learner is trained and predicts based on its experience over a time window of length $W$. Any online learning algorithm can be used as a base learner for this strategy. As shown in Algorithm \ref{alg_paired_SLMs} (Paired learning for OLMs), the stable learner is used for prediction, whereas the reactive learner is used to detect possible concept drifts:
\begin{itemize}[leftmargin=*]
	\item While the concept remains unchanged, the stable learner provides a better performance than the reactive learner (lines $6$ and $7$). 
	\item A counter is arranged to count the number of times within the window that the reactive learner predicts correctly the stream data, and the stable learner produces a bad prediction (lines $8$ to $10$). 
	\item If the proportion of these occurrences surpasses a threshold $\theta$ (line $11$), a concept change is assumed to have occurred, the stable learner is replaced with the reactive one, and the stable learner is reinitialized. 
	\item Finally, both stable and reactive learners are trained incrementally (lines $15$ and $16$). 
\end{itemize}
	
\resizebox{0.95\columnwidth}{!}{\begin{algorithm}[H]
	\DontPrintSemicolon
	\SetAlgoLined
	\SetKwInOut{Input}{Input}
	\SetKwInOut{Output}{Output}
	\Input{Preparatory data instances $\{(\textbf{X}_{t},y_t)\}_{t=0}^{t=P-1}$; training/testing data for the rest of the stream $\{(\textbf{X}_{t},y_{t})\}_{t=P}^{\infty}$; window size $W$ for the reactive learner; threshold $\theta$ for substituting the stable learner with the reactive one; stream learning model $f(\mathbf{X})$}
	\Output{Updated stable and reactive learners $f_{s}(\cdot)$ and $f_{r}(\cdot)$}
	Let $\mathbf{c}=[c^w]_{w=1}^W$ be a circular list of $W$ bits, each initially set to $c^w=0$ $\forall w$\;
	\For(\tcp*[h]{Preparatory process}){$t=0$ to $P-1$}{
		Train both $f_{s}(\cdot)$ and $f_{r}(\cdot)$ with $(\mathbf{X}_{t},y_{t})$\;
	}
	\For(\tcp*[h]{Stream learning}){$t=P$ to $\infty$}{
		Let $\widehat{y}_s=f_s(\mathbf{X}_t)$ (output of the model) and $\widehat{y}_r=f_r(\mathbf{X}_t)$\;
		Set $c^1=0$ and $c^w=c^{w-1}$ for $w=2,\ldots,W$\;
		\If{$\widehat{y}_{s} \neq y_{t}$ \textbf{and} $\widehat{y}_r = y_{t}$}{
			Set $c^1=1$\;
		}
		\If(\tcp*[h]{Drift detection}){$W^-1\sum_{w=1}^W c^w> \theta$}{
			Replace $f_s(\cdot)$ with $f_r(\cdot)$ (including its captured knowledge)\;
			Set $c^w=0$ $\forall w=1,\ldots,W$\;
		}
		Train $f_s(\dot)$ incrementally with $(\mathbf{X}_{t},y_{t})$\;
		Train $f_r(\cdot)$ from scratch over $\{(\mathbf{X}_{t'},y_{t'})\}_{t'=t-W+1}^t$\;		
	}
	\caption{Paired learning for OLMs}\label{alg_paired_SLMs}
\end{algorithm}}

Algorithm \ref{alg_carillon} (\texttt{\algname}) describes in detail the proposed algorithm. In essence the overall scheme relies on a couple of paired \texttt{sCA} learners: one in charge of making predictions on instances arriving in the data stream (\emph{stable} \texttt{sCA} learner), whereas the other (\emph{reactive} \texttt{sCA} learner) captures the most recent past knowledge in the stream (as reflected by a time window $W$). As previously explained for generic paired learning, the comparison between the accuracies scored by both learners over the time window permits to declare the presence of a concept drift when the reactive learner starts predicting more accurately than its stable counterpart. When this is the case, the grid state values of the reactive \texttt{sCA} is transferred to the stable automaton, and proceeds forward by predicting subsequent stream instances with an updated grid that potentially best represents the prevailing distribution of the stream. 

We now summarize the most interesting capabilities of the \texttt{\algname} algorithm:

\begin{algorithm}[h!]
	\DontPrintSemicolon
	\SetAlgoLined
	\SetKwInOut{Input}{Input}
	\SetKwInOut{Output}{Output}
	\Input{Preparatory data instances $\{(\textbf{X}_{t},y_{t})\}_{t=0}^{P-1}$; training/testing data for the rest of the stream $(\textbf{X}_{t},y_{t})_{t=P}^{\infty}$; window size $W$ for the reactive \texttt{sCA} learner; threshold $\theta$ for substituting the stable \texttt{sCA} learner with the reactive one; \texttt{sCA} parameters $G$, \smash{$f_{\lcirclearrowdown}(\cdot)$}, $f_{\boxplus}(\mathbf{c})$ and $R$ as per Algorithm \ref{alg_stream_automaton} }
	\Output{Updated stable and reactive learners $\texttt{sCA}_s$ and $\texttt{sCA}_{r}$}
	Let $\mathbf{c}=[c^w]_{w=1}^W$ be a circular list of $W$ bits, each initially set to $c^w=0$ $\forall w$\;
	Perform lines $1-28$ of Algorithm \ref{alg_stream_automaton} for $\texttt{sCA}_s$ and $\texttt{sCA}_r$ over $\{(\textbf{X}_{t},y_{t})\}_{t=0}^{P-1}$\;
	\For(\tcp*[h]{Stream learning}){$t=P$ to $\infty$}{
		Let $\widehat{y}_s=\texttt{sCA}_s(\mathbf{X}_t)$ and $\widehat{y}_r=\texttt{sCA}_r(\mathbf{X}_t)$ as per line 30 of Algorithm \ref{alg_stream_automaton}\;
		Set $c^1=0$ and $c^w=c^{w-1}$ for $w=2,\ldots,W$\;
		\If{$\widehat{y}_s \neq y_{t}$ and $\widehat{y}_r= y_{t}$}{
			Set $c^1=1$\;
		}
		\If(\tcp*[h]{Drift detection}){$W^-1\sum_{w=1}^W c^w> \theta$}{
			Copy grid state values of $\texttt{sCA}_r$ to $\texttt{sCA}_s$ (knowledge transfer)\;
			Clear $\texttt{sCA}_r$ and seed it with $\{(\mathbf{X}_{t'},y_{t'})\}_{t'=t-W+1}^t$\; Perform lines $1-28$ of Algorithm \ref{alg_stream_automaton} for $\texttt{sCA}_r$\;
			Set $c^w=0$ $\forall w=1,\ldots,W$\;
		}
	
		Train $\texttt{sCA}_s$ incrementally with $(\mathbf{X}_{t},y_{t})$\tcp*[h]{Lines $31-33$, Alg.\ref{alg_stream_automaton} }\;
		Train $\texttt{sCA}_r$ from scratch over $\{(\mathbf{X}_{t'},y_{t'})\}_{t'=t-W+1}^t$\;	
	}
	\caption{\texttt{\algname}}\label{alg_carillon}
\end{algorithm}

\begin{itemize}[leftmargin=*]
	\item \textit{Incremental learning}: as it relies on \texttt{sCA} learners, \texttt{\algname} shows a natural ability to learn incrementally, which is a crucial requirement under the constrained conditions present in RTA.
	\item \textit{Few parameters}: \texttt{\algname} requires only a few parameters, a warmly welcome characteristic that simplifies the parameter tuning process in drifting environments.
	\item \textit{Little data to be trained}: \texttt{\algname} can learn a data distribution with relatively few instances which, in the context of stream learning, involves a quicker preparation (warm-up) phase. However, depending on the grid size $G$, several generations may be required to make all cells be assigned a state. This is exemplified in Figure \ref{fig_few_samples_yellow}, which considers a dataset with $2$ continuous features $\mathbf{X}_t = (X_t^1,X_t^2)\in\mathtt{R}[0,1]$, and a binary target class $y_t\in\{0,1\}$ . Here, a $d\times G=2\times 5$ cellular automaton only needs $2$ iterations of the loops in lines 12 to 28 (hereafter referred to as \emph{generation}) until all cells are assigned one and only one state (see Algorithm \ref{alg_stream_automaton} ). However, the grid is coarse, and its distribution of states does not represent the distribution of the data stream. With a $d\times G = 2\times 10$ cellular automaton, $4$ generations are needed until no cells are left empty, and the representation of the data distribution is more finely grained. Finally, the $d\times G=2\times 20$ cellular automaton needs $7$ generations until no cells are left empty, yet the representation of the data distribution is well defined. A good balance between representativeness and complexity must be met in practice when deploying \texttt{\algname} in real setups.	
	\item \textit{Constrained scenarios}: in setups where computational costs must be reduced even more, a subsampling strategy is often recommended, where only a fraction of instances of the data stream is considered for model updating. In these scenarios, the capability of CA to represent the data distribution with a few instances fits perfectly. So does \texttt{\algname} by embracing \texttt{sCA} at its core.
	\item \textit{Stream learning under verification latency}: learning in an environment where labels do not become immediately available for model updating is known as \textit{verification latency}. This requires mechanisms and techniques to propagate class information forward through several time steps of unlabeled data. This remains as an open challenge in the stream learning community \citep*{openchallenges2014}. Here, \texttt{\algname} finds its gap as well, due to its capability of evolving through several generations and represents a data distribution from a few annotated data instances.
	\item \textit{Evolving scenarios}: by harnessing the concept of paired learners, \texttt{\algname} adapts to evolving conditions where changes (drifts) provoke that learning algorithms have to forget and learn the old and the new concept, respectively.
	\item \textit{Tractable and interpretable}: both characteristics lately targeted with particularly interest under the eXplainable Artificial Intelligence (XAI) paradigm \citep*{alej2019explainable}.
\end{itemize}

Finally, \texttt{\algname} enters the stream learning scene as a new base learner for the state-of-the-art.

\section{Experimental Setup}\label{exps}
We have designed two experiments in order to answer the three research questions posed in the Section \ref{intro}. The first experiment addresses the research questions \texttt{RQ1} and \texttt{RQ2} on the capability of \texttt{sCA} to learn incrementally and adapt to changes conditions, while the second experiment addresses \texttt{RQ3} by discussing on the competitiveness of our \texttt{\algname} algorithm with respect to other streaming learners.  

In all the experiments we have used the von Neumann neighborhood because it is linear in the number of dimensions of the instance space, so it scales well when dealing with problems of high dimensionality. In addition, this neighborhood is composed of less neighbors than in Moore's case, and the local rule is applied over less cells, which makes the process lighter and better for streaming scenarios in terms of computational and processing time costs. Regarding the performance measurement, we have adopted the so-called \textit{prequential accuracy} \citep*{dawid1999prequential} for all experiments, as it is a suitable metric to evaluate the learner performance in the presence of concept drift \citep*{dawid1999prequential,gama2013evaluating,gama2014survey}. This metric quantifies the average accuracy obtained by the prediction of each test instance before its learning in an online \textit{test-then-train} fashion, and is defined as:
\begin{equation}
preACC(t)\hspace{-1mm}=\hspace{-1mm}\left\lbrace
\hspace{-2mm}\begin{array}{ll}
preACC_{ex}(t), & \hspace{-2mm}\mbox{if $t= t_{ref}$,} \\
preACC_{ex}(t\mbox{-}1) + \dfrac{preACC_{ex}(t)-preACC_{ex}(t\mbox{-}1)}{t - t_{ref} + 1}, & \hspace{-2mm}\mbox{otherwise,}
\end{array}\right.
\end{equation}
where $preACC_{ex}(t)=0$ if the prediction of the test instance at time $t$ before its learning is wrong and $1$ if it is correct; and $t_{ref}$ is a reference time that fixes the first time step used in the calculation. This reference time allows isolating the computation of the prequential accuracy before and after a drift has started.

As for the initialization of \texttt{sCA}, following Algorithm \ref{alg_stream_automaton} we have used a reduced group of $P$ preparatory instances. In the second experiment we have also used these instances to carry out the hyper-parameter tuning of the OLMs under analysis. The selection of the size of this preparatory data usually depends on the available memory or the processing time we can take to collect or process these data. Finally, following the recommendations of \citep*{fawcett2008data}, we have assigned one grid dimension to each attribute.

%Finally, the availability of the source code for the algorithms and experiments is detailed in Section \ref{code}.

\subsection{First Experiment: Addressing \texttt{RQ1} and \texttt{RQ2} with \texttt{sCA}}\label{synt_dat}

In order to address \texttt{RQ1} and \texttt{RQ2}, we have used several two-dimensional synthetic datasets for the first experiment, where results can be easily visualised and interpreted. Synthetic data are advisable because in real datasets is not possible to know exactly when a drift appears, which type of drift emerges, or even if there is any confirmed drift. Thus, it is not possible to analyze the behavior of \texttt{sCA} in the presence of concept drift by only relying on real-world datasets. 

This being said, for this experiment we have used the \texttt{sCA} detailed in Section \ref{stream_CA} (Algorithm \ref{alg_stream_automaton} ). We have compared its naive version with the one using an active adaptation mechanism: assuming a perfect drift detection (the drift point is known beforehand), once a drift occurs the grid of the \texttt{sCA} model is seeded with a $W$-sized window of past instances, which allows representing the current concept (data distribution). Then, \texttt{sCA} progresses again through several generations (and applying the local rule) until all cells are assigned a state. As highlighted before, this ability of CA to represent a data distribution from a few instances is very valuable here, where only a limited sliding window is used to initialize the \texttt{sCA} grid. This experiment is designed to illustrate and visualize the operation of \texttt{sCA} in two dimensions (two features).

We use the renowned set of four balanced synthetic datasets (\texttt{CIRCLE}, \texttt{LINE}, \texttt{SINEH}, and \texttt{SINEV}) described in \citep*{minku2009impact}. Very briefly, these datasets contain one simulated drift characterized by low and high speed, resulting $2$ different types of drift for each dataset. Speed is the inverse of the time taken for a new concept to completely replace the previous one. Each dataset has $2,000$ instances ($t\in\{1,\ldots,2,000\}$), $2$ normalized ($[0,1]$) continuous features $\mathbf{X}_t = (X_t^1,X_t^2)$, and a binary target class $y_t\in\{0,1\}$. Drift occurs at $t=1,000$ (where it goes from the old concept to the new one), and the drifting period is $1$ for abrupt drifts (high speed datasets) and $500$ for gradual ones (low speed datasets). For abrupt drifts we have opted for a small window size of $W=25$, whereas for gradual drifts we need a bigger window of $W=100$ \citep*{gama2013evaluating}. The size of this window will depend on the type of drift: a small window can assure fast adaptability in abrupt changes, while a large window produces lower variance estimators in stable phases, but cannot react quickly to gradual changes. Thus, sliding windows is a very common form of the so-called \textit{forgetting mechanism} \citep*{gama2010knowledge}, where outdated data is discarded and adapt to the most recent state of the nature, then emphasis is placed on error calculation from the most recent data instances. 

\subsection{Second Experiment: Addressing \texttt{RQ3} with \texttt{\algname}}\label{real_dat}

For the second experiment we have resorted to real-world datasets to confirm the competitive performance of \texttt{\algname} in a comparison with recognized OLMs in the literature, answering \texttt{RQ3}. In this case, since we deal with problems having $n$ features, an extensive use is to assign one grid dimension of the CA to each feature of the dataset. In this experiment, \texttt{\algname} is compared with paired learning models (see Algorithm \ref{alg_paired_SLMs} ) using at their core different online learning algorithms from the literature. These comparison counterparts have been selected over others due to the fact that they are well-established methods in the stream learning community. Also because their implementations are reliable and easily accessible in well-known Python frameworks such as scikit-multiflow \citep*{montiel2018scikit} and scikit-learn \citep*{pedregosa2011scikit}. As this is the first experiment carried out to know about their performance as incremental learners with drift adaptation, we have selected these reputed methods:
\begin{itemize}[leftmargin=*]
	\item \textit{Stochastic Gradient Descent Classifier} (\texttt{SGDC}) \citep*{bottou2010large} implements an stochastic gradient descent learning algorithm which supports different loss functions and penalties for classification.
	\item \textit{Hoeffding Tree} (\texttt{HTC}) \citep*{domingos2000mining}, also known as Very Fast Decision Tree (VFDT), is an incremental anytime decision tree induction algorithm capable of learning from massive data streams, assuming that the distribution of data instances does not change over time, and exploiting the fact that a small instance can often be enough to choose an optimal splitting attribute.
	\item \textit{Passive Agressive Classifier} (\texttt{PAC}) \citep*{crammer2006online} focuses on the target variable of linear regression functions, $\hat{y_{t}}=\bm{\omega}_{t}^{T} \cdot \textbf{X}_{t}$, where $\bm{\omega}_{t}$ is the incrementally learned weight vector. After a prediction is made, the algorithm receives the true target value $y_{t}$ and an instantaneous $\varepsilon$-insensitive hinge loss function is computed to update the weight vector. This loss function was specifically designed to work with stream data, and is analogous to a standard hinge loss. The role of $\varepsilon$ is to allow for a lower tolerance of prediction errors. Then, when a round finalizes, the algorithm uses $\bm{\omega}_{t}$ and the instance $(\textbf{X}_{t},y_{t})$ to produce a new weight vector $\bm{\omega}_{t+1}$, which is then used to predict the next incoming instance.   
	\item\textit{k-Nearest Neighbours} (\texttt{KNN}) \citep*{read2012batch}, a well-known lazy learner where the output is given by the labels of the $K$ training instances closest (under a certain distance measure) to the query instance $\mathbf{X}_t$. In its \textit{streamified} version, it works by keeping track of a sliding window of past training instances. Whenever a query request is executed, the algorithm will search over its stored instances and find the $K\leq W$ closest neighbors in the window, again as per the selected distance metric.
\end{itemize}

Three real-world datasets have been utilized in this second experiment:
\begin{itemize}[leftmargin=*]
	\item The Australian New South Wales Electricity Market dataset (\texttt{ELEC2}) \citep*{gama2004learning} is a widely adopted real-world dataset in studies related to stream processing/mining, particularly in those focused on non-stationary settings. In essence the dataset represents electricity prices over time, which are not fixed and become affected by the demand and supply dynamics. It contains $45,312$ instances dated from $7$ May $1996$ to $5$ December $1998$. Each instance $\mathbf{X}_t$ of the dataset refers to a period of $30$ minutes, and has $d=5$ features (day of week, time stamp, NSW electricity demand, Vic electricity demand, and scheduled electricity transfer between states). The target variable $y_t$ to be predicted is binary, and identifies the change of the price related to a moving average of the last $24$ hours. The class level only reflect deviations of the price on a one day average, and removes the impact of longer term price trends.
	\item The Give Me Some Credit dataset (\texttt{GMSC}\footnote{Available at: \url{https://www.kaggle.com/c/GiveMeSomeCredit}. Last accessed on December 5th, 2019.}) correspond to a credit scoring task in which the goal is to decide whether a loan should be granted or not (binary target $y_t$). This is a core decision for banks due to the risk of unexpected expenses and future lawsuits. The dataset comprises supervised historical data of $150,000$ borrowers described by $d=10$ features.
	\item The \texttt{POKER-HAND} dataset\footnote{Available at: \url{https://archive.ics.uci.edu/ml/datasets/Poker+Hand}. Last accessed on December 5th, 2019.} originally consists of $1,000,000$ stream instances. Each record is an example of a hand consisting of five playing cards drawn from a standard deck of $52$ cards. Each card is described by $2$ attributes (suit and rank), giving rise to a total of $d=10$ predictive attributes for every stream instance $\mathbf{X}_t$. The goal is to predict the poker hand upon such features, out of a set of $|\mathcal{S}|=10$ possible poker hands (i.e., multiclass classification task).
\end{itemize}

In order to alleviate the computational effort needed to run the entire benchmark, we have selected $20,000$ instances of each dataset. We have considered the first $50\%$ of the datasets ($10,000$ instances) to tune the parameters of the OLMs and to seed \texttt{\algname}, whereas the rest $50\%$ has been used for prediction and performance assessment. Each experiment has been carried out $25$ times in the case of \texttt{SGDC}, \texttt{PAC} and \texttt{KNNC} to account for the stochasticity of the hyperparameter search algorithm and their learning procedure. Indeed, the considered online models are sensible to parameter values. Thus we have performed a randomized hyper-parameter tuning over preparatory instances, obtaining different hyper parameter settings for each run. Table \ref{tab_parameters_conf} summarizes the parameters that have participated in each randomized search process, following the nomenclature of the scikit-learn\footnote{https://scikit-learn.org/. Last access in December 5th, 2019.} and scikit-multiflow\footnote{https://scikit-multiflow.github.io/. Last access in December 5th, 2019.} frameworks. In the case of \texttt{HTC}, we found experimentally that its default values (see documentation of scikit-multiflow for more details) have worked very well in all datasets, so it was not necessary to carry out the randomized search on hyper parameters, which helped speed up the experimentation. Finally, the same window size $W$ than \texttt{\algname} has been assigned to \texttt{KNNC} for the sake of a fair comparison between them. 
\begin{table}[h!]
	\centering
	\resizebox{0.8\textwidth}{!}{%
		\begin{tabular}{@{}cccc@{}}
			\toprule
			\textbf{Model} & \textbf{Parameters} & \textbf{Values used} & \textbf{Tuning} \\ \midrule
			\multirow{6}{*}{\texttt{SGDC}} & \textit{alpha} & $10^{-x}$, $x\in\{-1,-2,\ldots,-6\}$ & \multirow{6}{*}{yes} \\
			& \textit{loss} & `perceptron', `hinge', `log', `modified\_huber', `squared\_hinge'\\
			& \textit{learning\_rate} & `constant', `optimal', `invscaling', `adaptive' \\
			& \textit{eta0} & $0.1,0.5,1.0$ \\
			& \textit{penalty} & None, `l2', `l1', `elasticnet' \\
			& \textit{max\_iter} & $1,100,200,500$ \\ \cmidrule(l){1-4}
			\multirow{2}{*}{\texttt{PAC}} & \textit{C} & $0.001, 0.005, 0.01, 0.05, 0.1, 0.5, 1.0$ & \multirow{2}{*}{yes} \\ 
			& \textit{max\_iter} & $1,100,200,500$  \\ \cmidrule(l){1-4}
			\multirow{4}{*}{\texttt{KNNC}} & \textit{n\_neighbors} & $5,10,15,25,50$ & \multirow{4}{*}{yes} \\
			& \textit{leaf\_size} & $5, 10,20,30$  \\
			& \textit{algorithm} & `auto'  \\
			& \textit{weights} & `uniform', `distance'  \\ \cmidrule(l){1-4}			
			\multirow{4}{*}{\texttt{HTC}} & \textit{grace\_period} & $200$ & \multirow{4}{*}{no} \\
			& \textit{split\_criterion} & `info\_gain' \\
			& \textit{leaf\_prediction} & `nba' \\
			& \textit{nb\_threshold} & $0$ \\ 
			
			\bottomrule
		\end{tabular}%
	}
	\caption{Parameters that have participated in each randomized searching process of \texttt{SGDC}, \texttt{PAC} and \texttt{KNNC}. In the case of \texttt{HTC} its default values show a good enough performance and has not participated in the randomized hyperparameter searching process. This has alleviated the computing cost of the experiments.}
	\label{tab_parameters_conf}
\end{table}

Finally, Table \ref{tab_carillon_parameters} summarizes the parameter settings of \texttt{\algname} and the OLMs used for the real-world experiments. 
\begin{table}[h!]
	\centering
	\resizebox{0.95\textwidth}{!}{%
		\begin{tabular}{cccccc}
			\toprule
			\multirow{2}{*}{\textbf{Model}} & \multirow{2}{*}{\textbf{Parameter}} & \multicolumn{3}{c}{\textbf{Real data}} & \multirow{2}{*}{\textbf{Synthetic data}} \\
			\cmidrule{3-5} 
			& & \texttt{ELEC2} & \texttt{GMSC} & \texttt{POKER-HAND} &  \\ 
			\midrule
			\multirow{2}{*}{All models} & $P$ & \multicolumn{3}{c}{$50\%$ of the considered stream length} & $5\%$ of the considered stream length \\
			& $W$ & $50$ instances & $250$ instances & $250$ instances & $25,50,100$ instances \\
			\midrule
			\multirow{6}{*}{\rotatebox{90}{\texttt{\algname}}} & $f_{\boxplus}(\cdot)$ & \multicolumn{4}{c}{\Vhrulefill von Neumann\Vhrulefill} \\			
			& $f_{\lcirclearrowdown}(\cdot)$ & \multicolumn{4}{c}{\Vhrulefill Majority voting from Expression \eqref{eq_majvot} \Vhrulefill} \\
			& $R$ & \multicolumn{4}{c}{\Vhrulefill 1 ({\footnotesize \smash{\PlusCenterOpen}})\Vhrulefill} \\
			& $\mathcal{S}$ & $\{0,1\}$ & $\{0,1\}$ & $\{0,1,\ldots,9\}$ & $\{0,1\}$\\		
			& $d\times G$ & $5\times5$ & $10\times3$ & $10\times3$ & $2\times5$, $2\times10$, $2\times20$\\
			& $\theta$ & $0.0$5 & $0.01$ & $0.001$ & Not applicable \\ 
			\midrule
			\texttt{SGDC} & & $0.8$ & $0.001$ & $0.1$ & \\
			\texttt{HTC} & & $0.1$ & $0.0001$ & $0.01$ & \\
			\texttt{PAC} & $\theta$ & $0.8$ & $0.001$ & $0.1$ & Not used \\
			\texttt{KNNC} & & $0.1$ & $0.001$ & $0.001$ & \\
			\bottomrule
		\end{tabular}}
	\caption{Parameters configuration for \algname and the considered OLMs in synthetic (\texttt{RQ1} and \texttt{RQ2}) and real-world experiments (\texttt{RQ3}).}
	\label{tab_carillon_parameters}
\end{table}

\section{Results and Analysis}\label{reses}

Next, we present the results for the two experiments, organizing the foregoing analysis and discussion in terms of the research questions posed in Section \ref{intro}:

\subsection{\texttt{RQ1}: Does \texttt{sCA} act as a real incremental learner?}

\begin{figure}[h!]
	\begin{subfigure}{\textwidth}
		\centering
		% include first image		
		\includegraphics[width=0.85\linewidth]{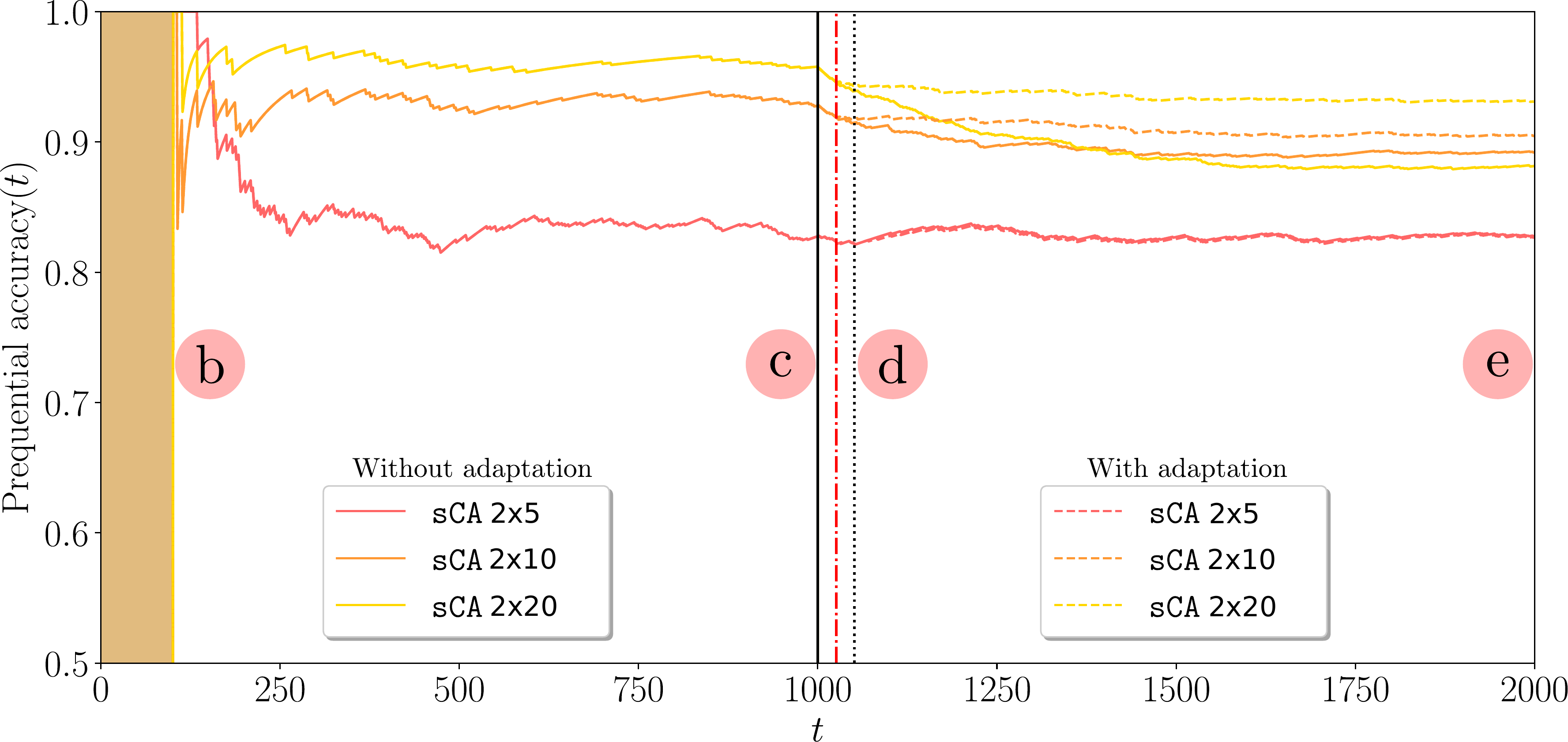} 
		\caption{}
		\label{fig_circleG31_graph}
	\end{subfigure}
	\begin{subfigure}{.5\textwidth}
		\centering
		% include first image		
		\includegraphics[width=\linewidth]{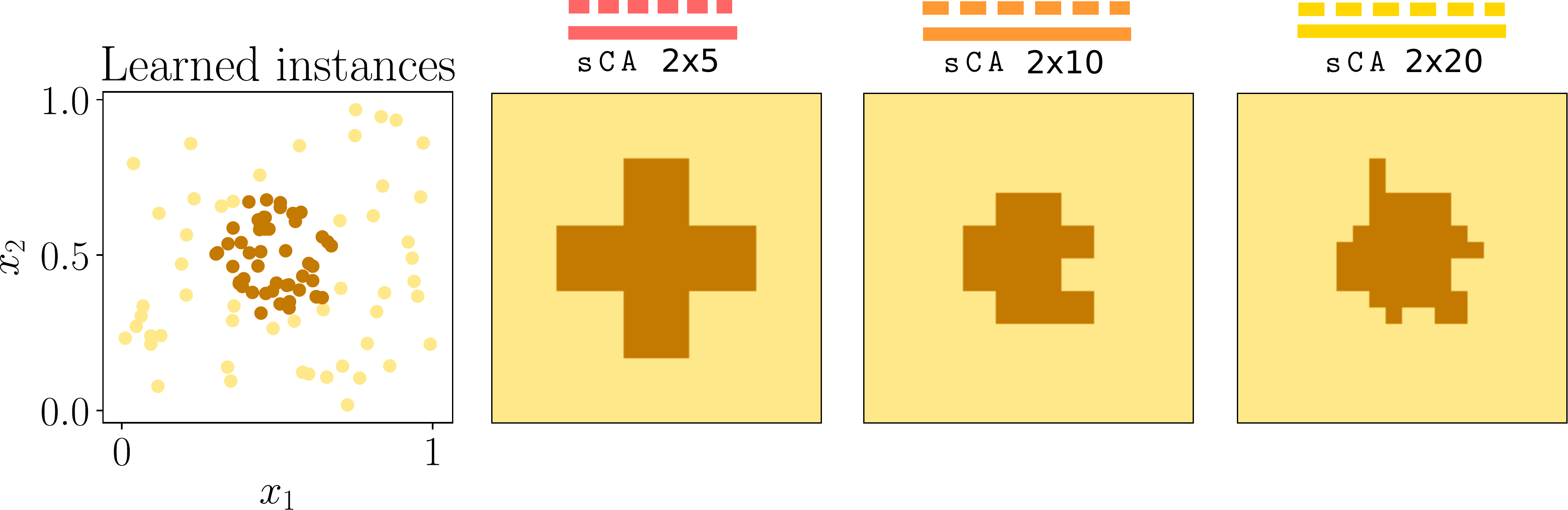} 
		\caption{}
		\label{fig_circleG31_CAs_1}
	\end{subfigure}
	\begin{subfigure}{.5\textwidth}
		\centering
		% include first image		
		\includegraphics[width=\linewidth]{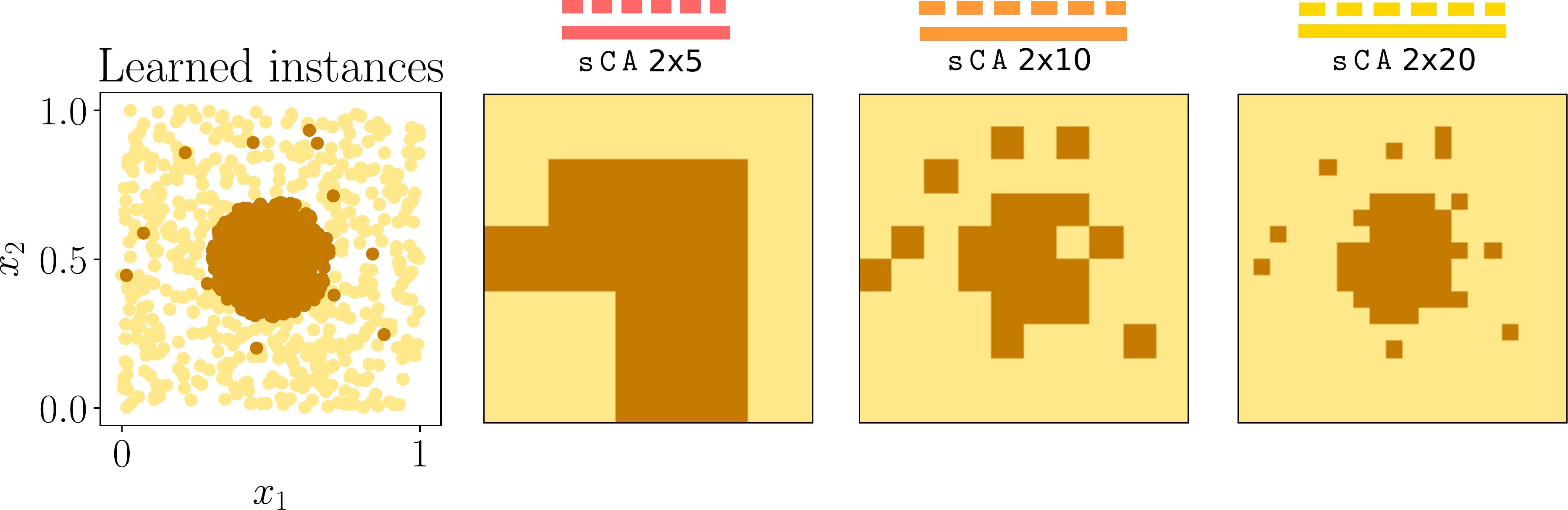} 
		\caption{}
		\label{fig_circleG31_CAs_2}
	\end{subfigure}
	\begin{subfigure}{0.5\textwidth}
		\centering
		% include first image		
		\includegraphics[width=\linewidth]{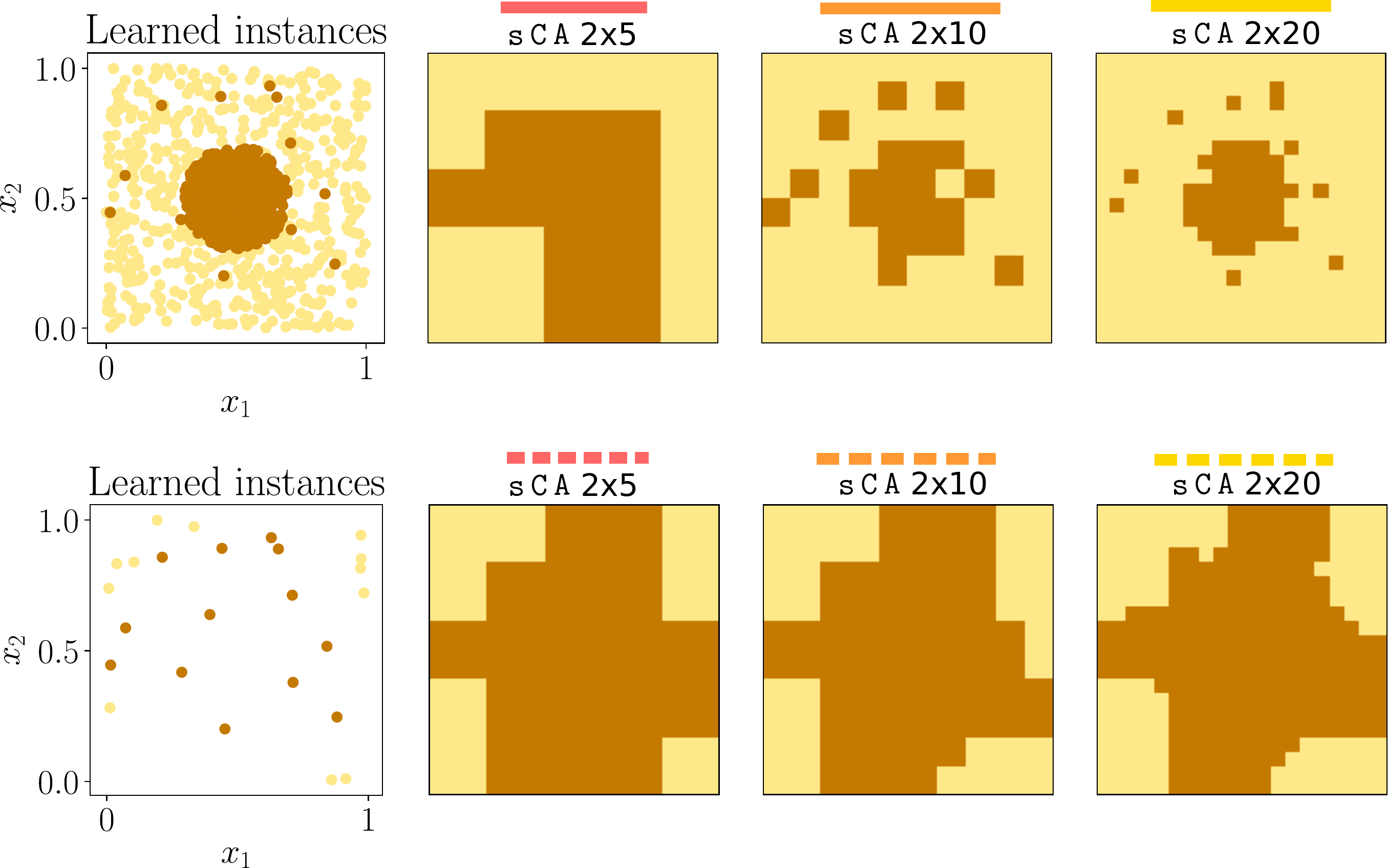} 
		\caption{}
		\label{fig_circleG31_CAs_3}
	\end{subfigure}
	\begin{subfigure}{0.5\textwidth}
		\centering
		% include first image		
		\includegraphics[width=\linewidth]{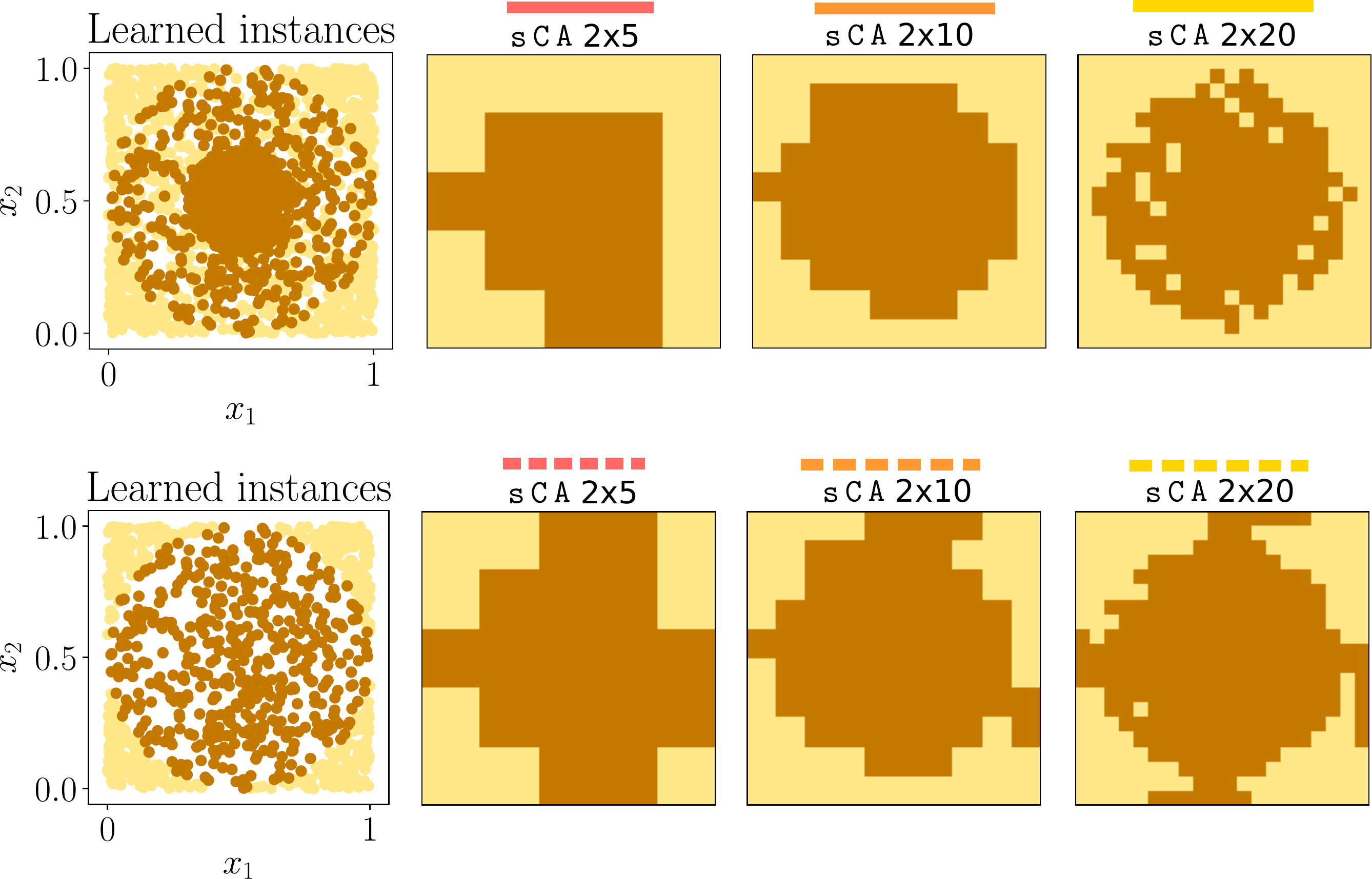} 
		\caption{}
		\label{fig_circleG31_CAs_4}
	\end{subfigure}
	\caption{Learning and adaptation of von Neumann's \texttt{sCA} of different grid sizes for the \texttt{CIRCLE} dataset, which exhibits an abrupt drift at $t=1000$. The drift is detected at $t=1025$ and then the adaptation mechanism, which uses a window of instances $W=25$, is triggered. (a) Performance comparison at points \smash{\circleddefaults{b}} (initial), \smash{\circleddefaults{c}} (just before the drift occurs), \smash{\circleddefaults{d}} ($W=25$ instances after drift detection and the adaptation mechanism were triggered), and \smash{\circleddefaults{e}} (final). The dashdotted line points out the drift detection and the dotted line the point in which the performance is measured. The shaded area corresponds to the preparatory instances. (b) The cells of \texttt{sCA}s are seeded with the preparatory instances (\smash{\circleddefaults{b}}). (c) The learning process of \texttt{sCA} before drift occurs (\smash{\circleddefaults{c}}). (d) The learning process of \texttt{sCA} after the drift occurs, and how those with the adaptation mechanism is initialized and seeded with a set of $W=25$ past instances (\smash{\circleddefaults{d}}). (e) The learning process of \texttt{sCA} until the end of the stream (\smash{\circleddefaults{e}}).}
	\label{fig_synths_circle31}
\end{figure}

\begin{figure}[h!]
	\begin{subfigure}{\textwidth}
		\centering
		% include first image		
		\includegraphics[width=0.85\linewidth]{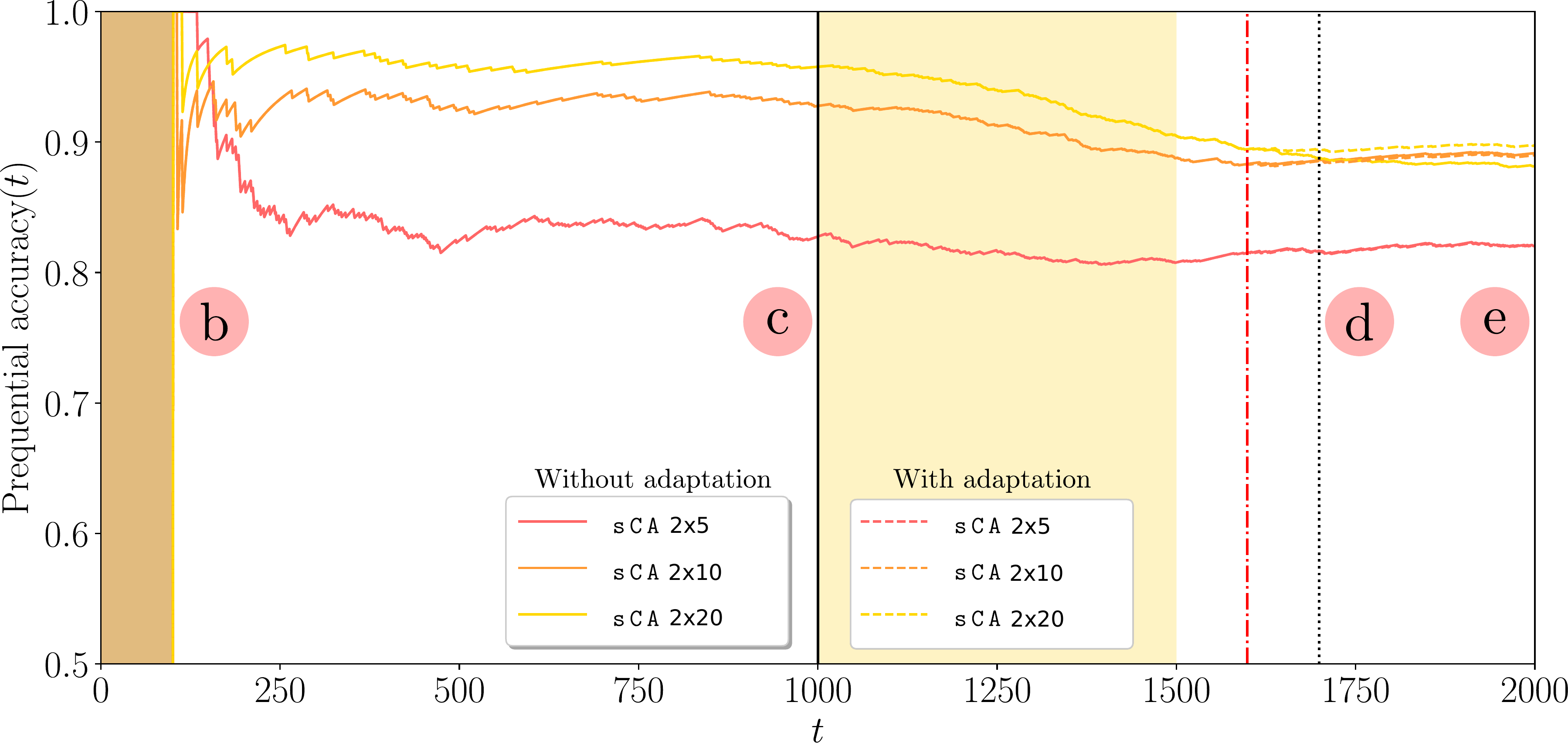} 
		\caption{}
		\label{fig_circleG33_graph}
	\end{subfigure}
	\begin{subfigure}{.5\textwidth}
		\centering
		% include first image		
		\includegraphics[width=\linewidth]{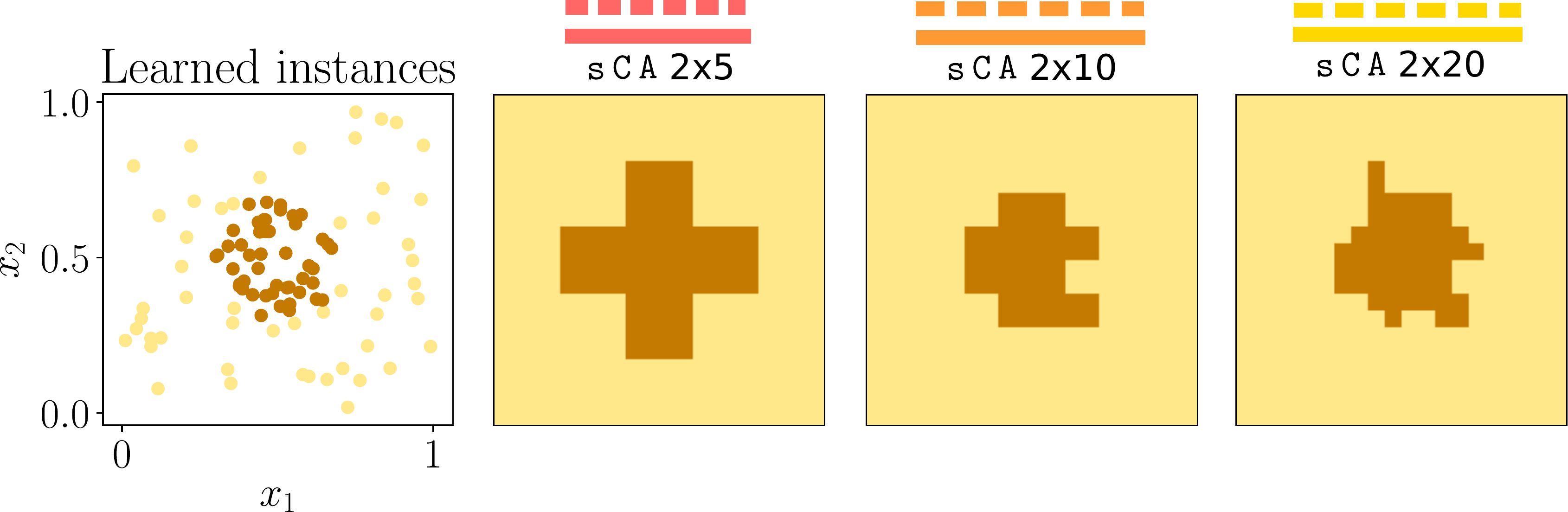} 
		\caption{}
		\label{fig_circleG33_CAs_1}
	\end{subfigure}
	\begin{subfigure}{.5\textwidth}
		\centering
		% include first image		
		\includegraphics[width=\linewidth]{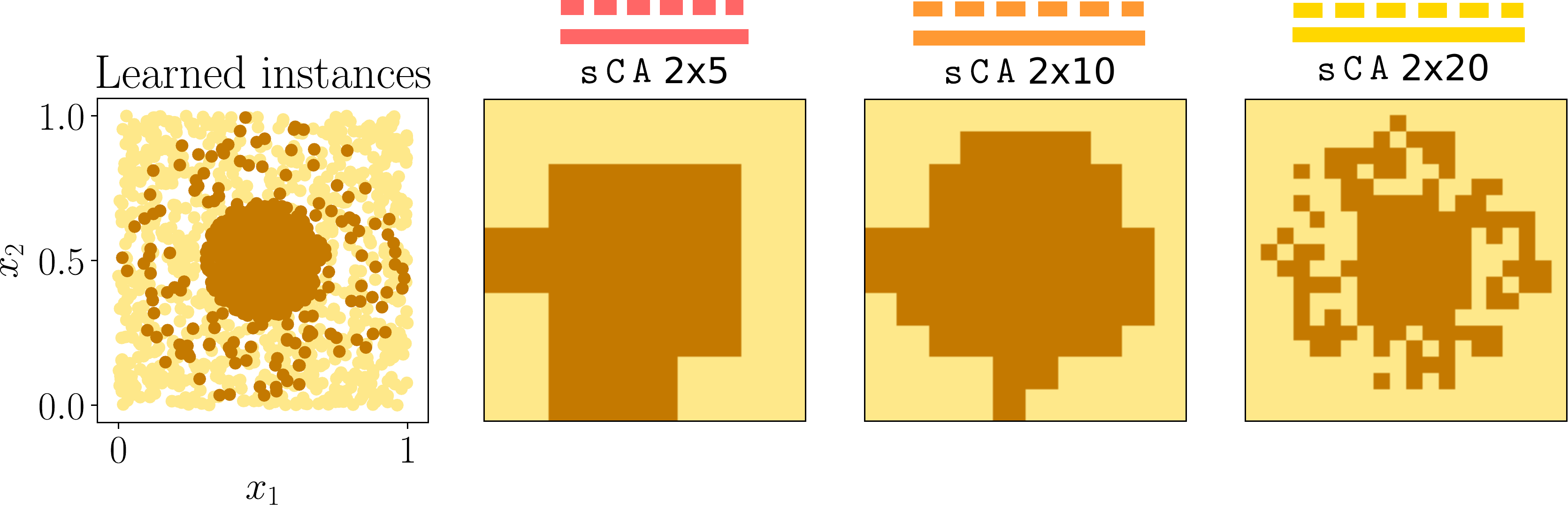} 
		\caption{}
		\label{fig_circleG33_CAs_2}
	\end{subfigure}
	\begin{subfigure}{0.5\textwidth}
		\centering
		% include first image		
		\includegraphics[width=\linewidth]{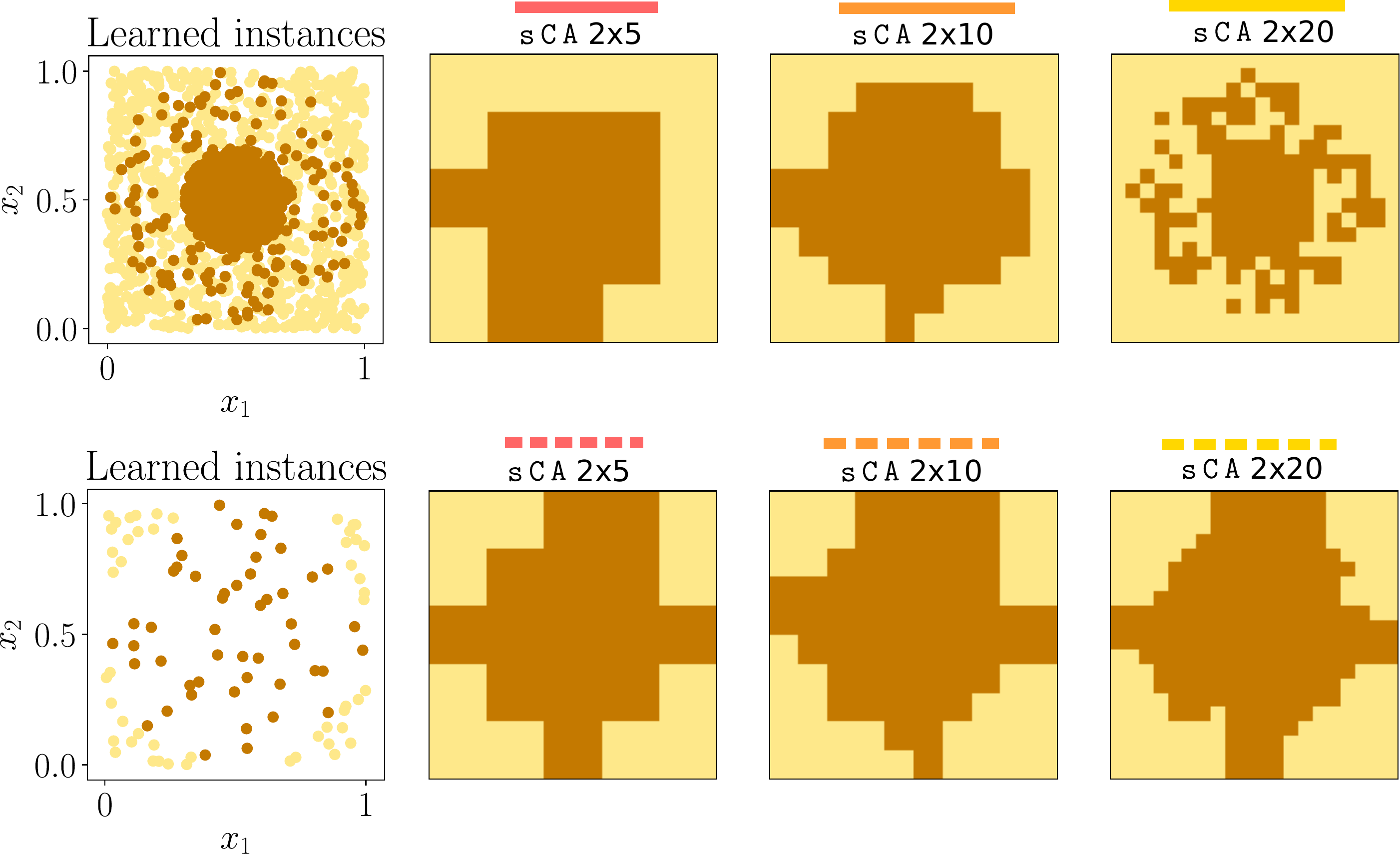} 
		\caption{}
		\label{fig_circleG33_CAs_3}
	\end{subfigure}
	\begin{subfigure}{0.5\textwidth}
		\centering
		% include first image		
		\includegraphics[width=\linewidth]{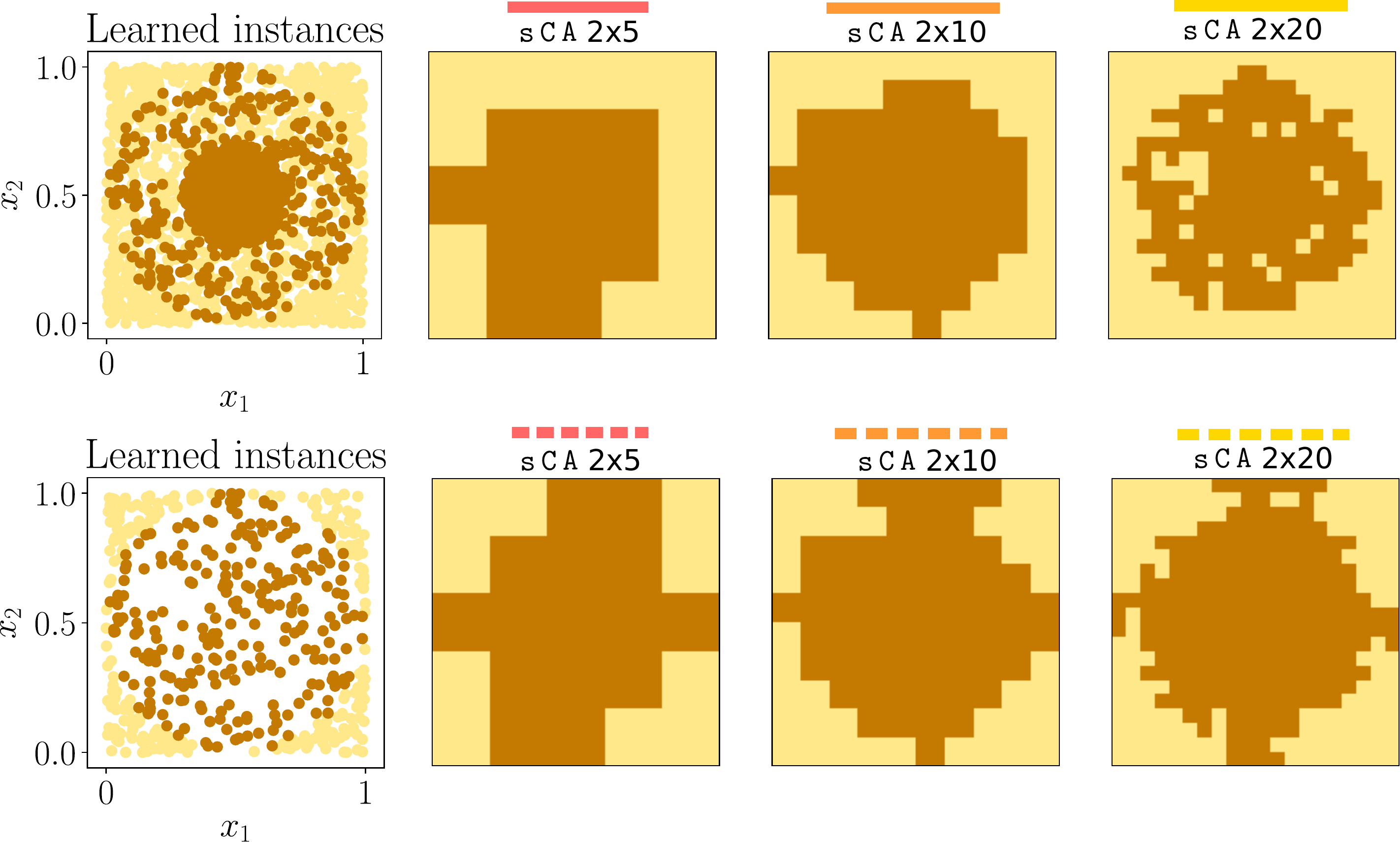} 
		\caption{}
		\label{fig_circleG33_CAs_4}
	\end{subfigure}
	\caption{Learning and adaptation of von Neumann's \texttt{sCA} of different grid sizes for the \texttt{CIRCLE} dataset, which exhibits a gradual drift at $t=1000$. The drift is detected at $t=1600$, and then the adaptation mechanism, which uses a window of instances $W=100$, is triggered. The interpretation of the remaining plots is the same than those in Figure \ref{fig_synths_circle31}.}
	\label{fig_synths_circle33}
\end{figure}

\begin{figure}[h!]
	\begin{subfigure}{\textwidth}
		\centering
		% include first image		
		\includegraphics[width=0.85\linewidth]{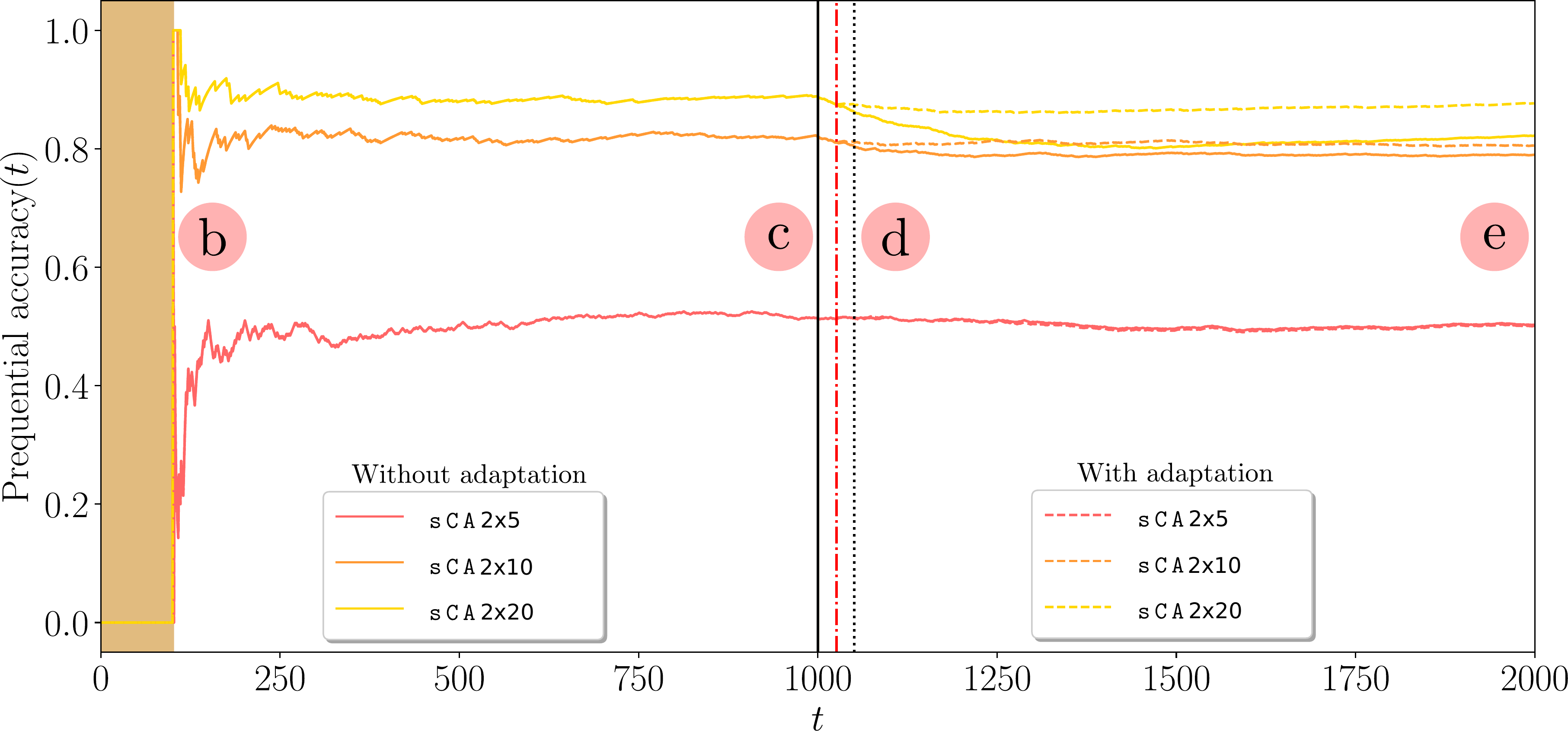} 
		\caption{}
		\label{fig_sineH31_graph}
	\end{subfigure}
	\begin{subfigure}{.5\textwidth}
		\centering
		% include first image		
		\includegraphics[width=\linewidth]{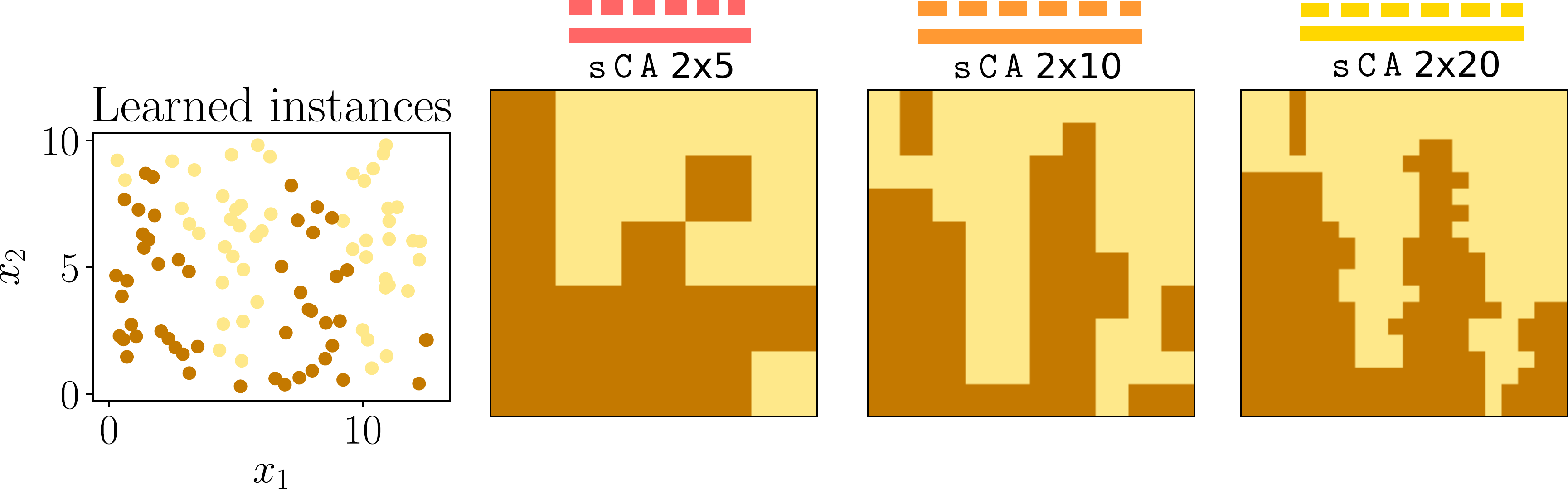} 
		\caption{}
		\label{fig_sineH31_CAs_1}
	\end{subfigure}
	\begin{subfigure}{.5\textwidth}
		\centering
		% include first image		
		\includegraphics[width=\linewidth]{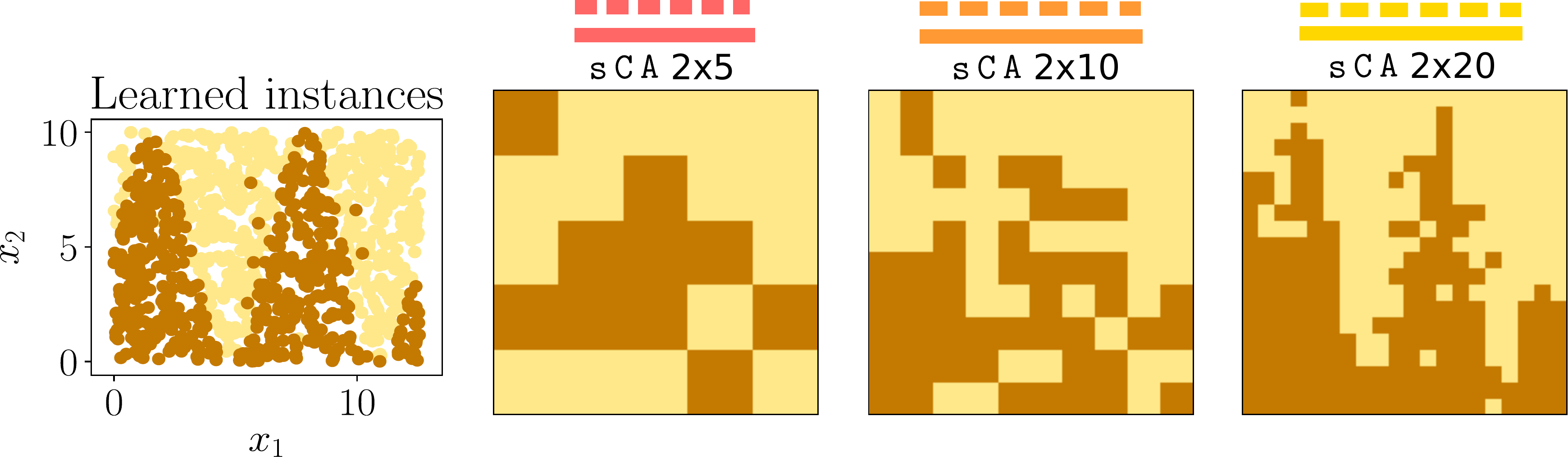} 
		\caption{}
		\label{fig_sineH31_CAs_2}
	\end{subfigure}
	\begin{subfigure}{0.5\textwidth}
		\centering
		% include first image		
		\includegraphics[width=\linewidth]{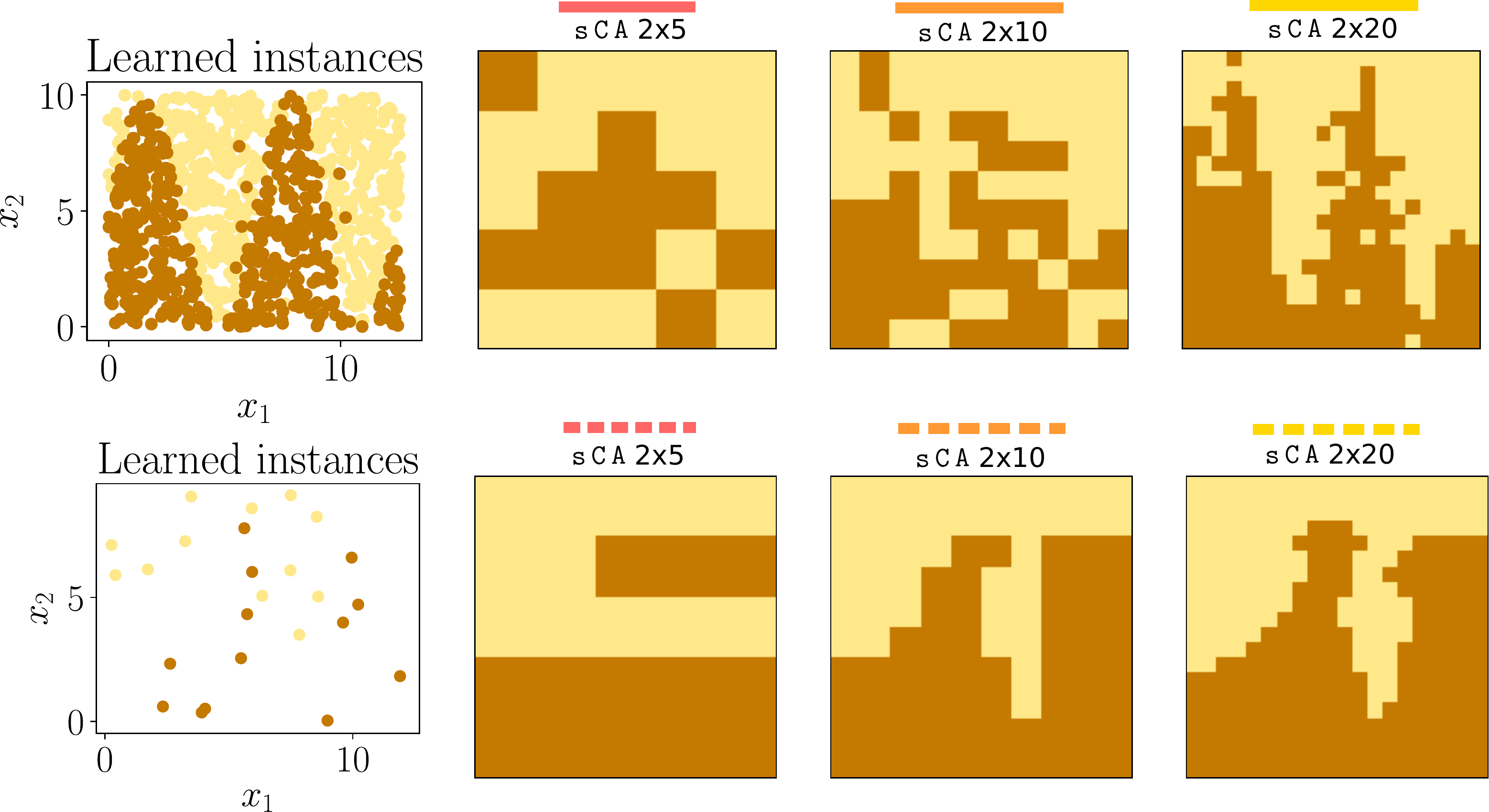} 
		\caption{}
		\label{fig_sineH31_CAs_3}
	\end{subfigure}
	\begin{subfigure}{0.5\textwidth}
		\centering
		% include first image		
		\includegraphics[width=\linewidth]{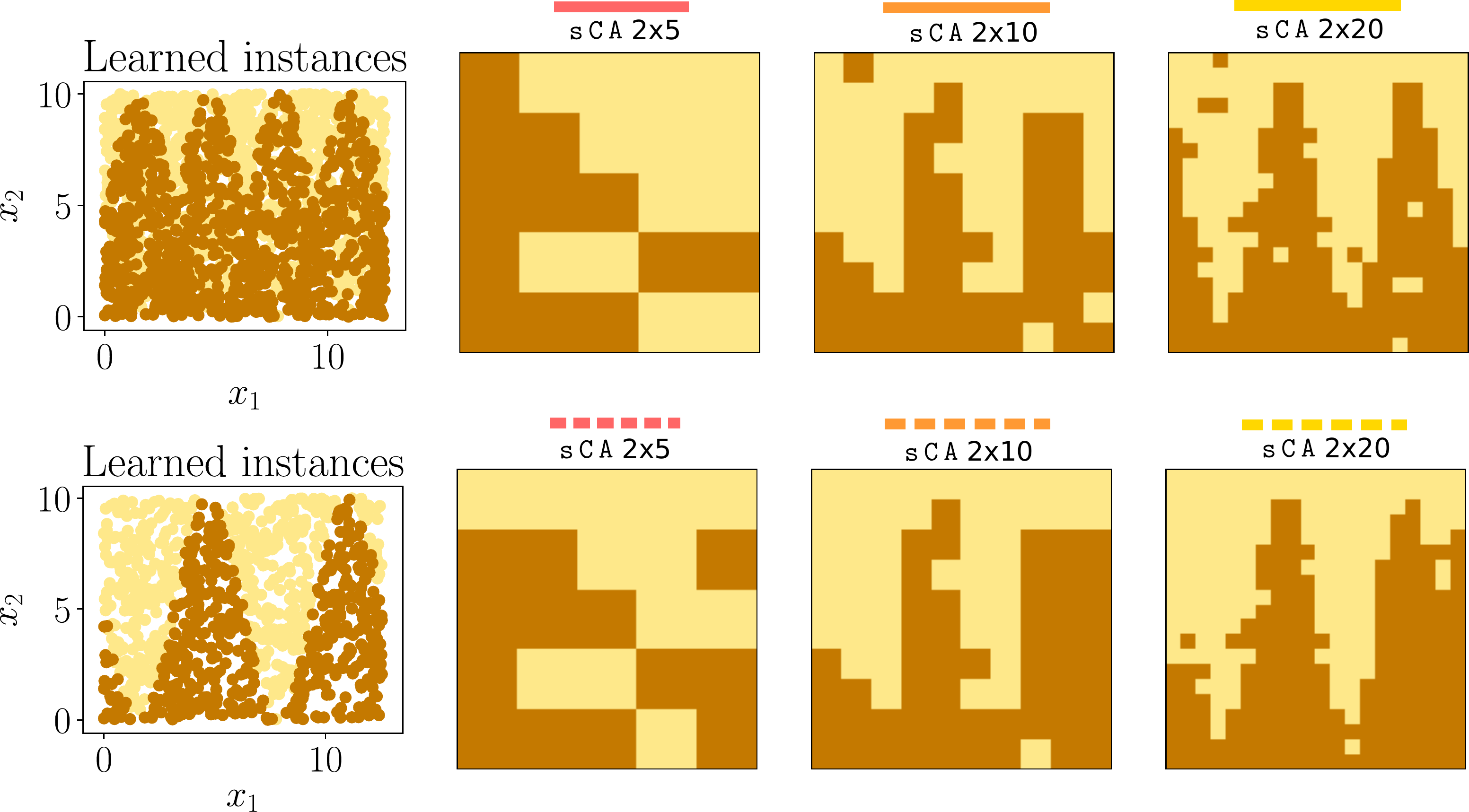} 
		\caption{}
		\label{fig_sineH31_CAs_4}
	\end{subfigure}
	\caption{Learning and adaptation of von Neumann's \texttt{sCA} of different grid sizes for the \texttt{SINEH} dataset, which exhibits an abrupt drift at $t=1000$. The drift is detected at $t=1025$ and then the adaptation mechanism, which uses a window of instances $W=25$, is triggered. The interpretation of the remaining plots is the same than those in Figure \ref{fig_synths_circle31}.}
	\label{fig_synths_sineH31}
\end{figure}

\begin{figure}[h!]
	\begin{subfigure}{\textwidth}
		\centering
		% include first image		
		\includegraphics[width=0.85\linewidth]{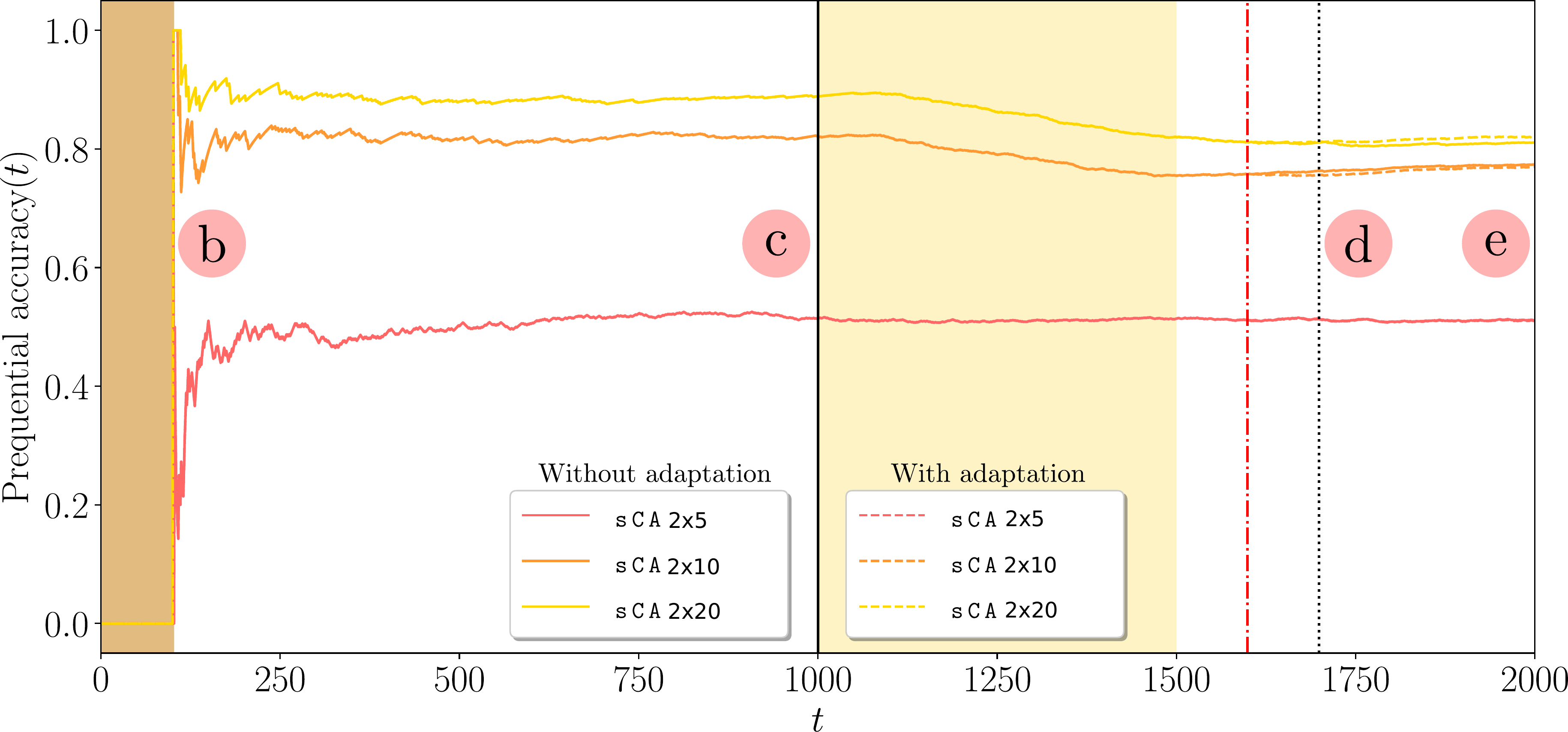} 
		\caption{}
		\label{fig_sineH33_graph}
	\end{subfigure}
	\begin{subfigure}{.5\textwidth}
		\centering
		% include first image		
		\includegraphics[width=\linewidth]{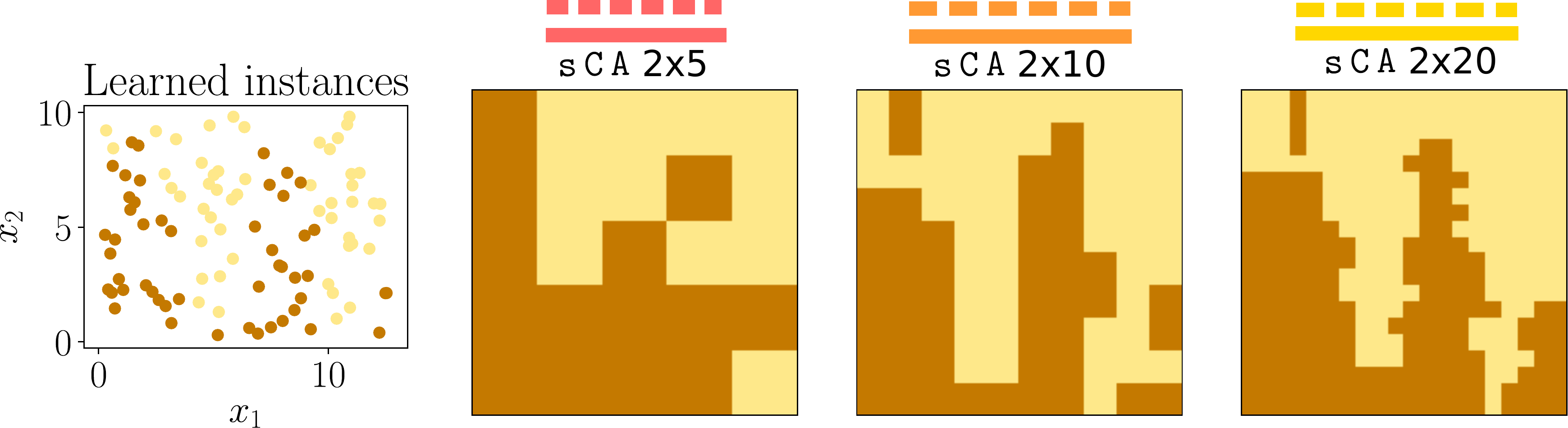} 
		\caption{}
		\label{fig_sineH33_CAs_1}
	\end{subfigure}
	\begin{subfigure}{.5\textwidth}
		\centering
		% include first image		
		\includegraphics[width=\linewidth]{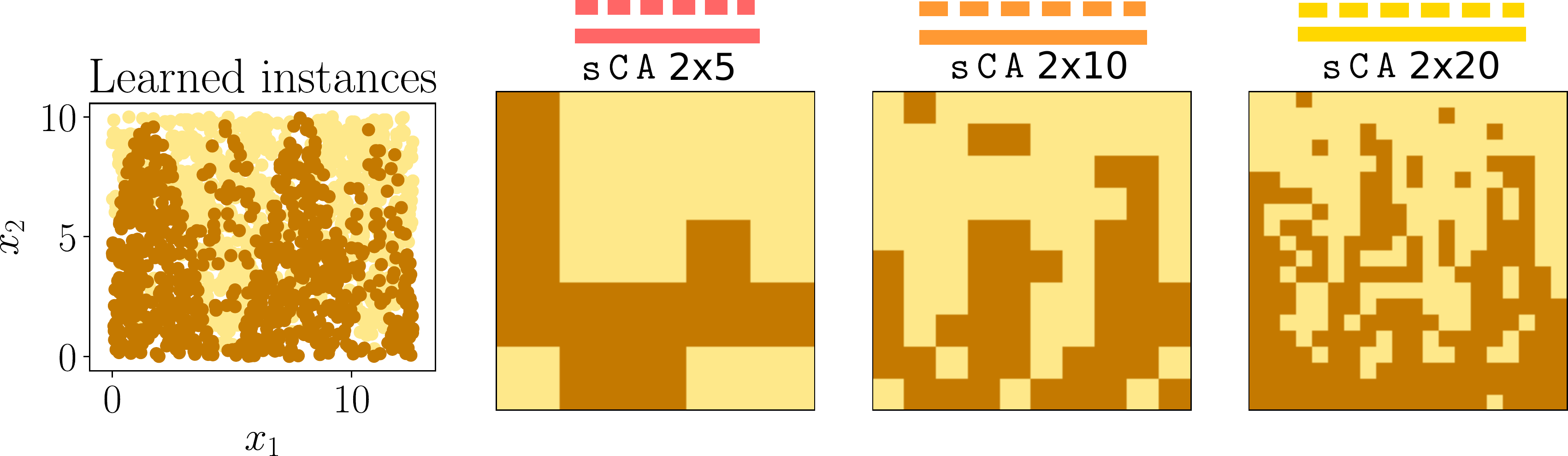} 
		\caption{}
		\label{fig_sineH33_CAs_2}
	\end{subfigure}
	\begin{subfigure}{0.5\textwidth}
		\centering
		% include first image		
		\includegraphics[width=\linewidth]{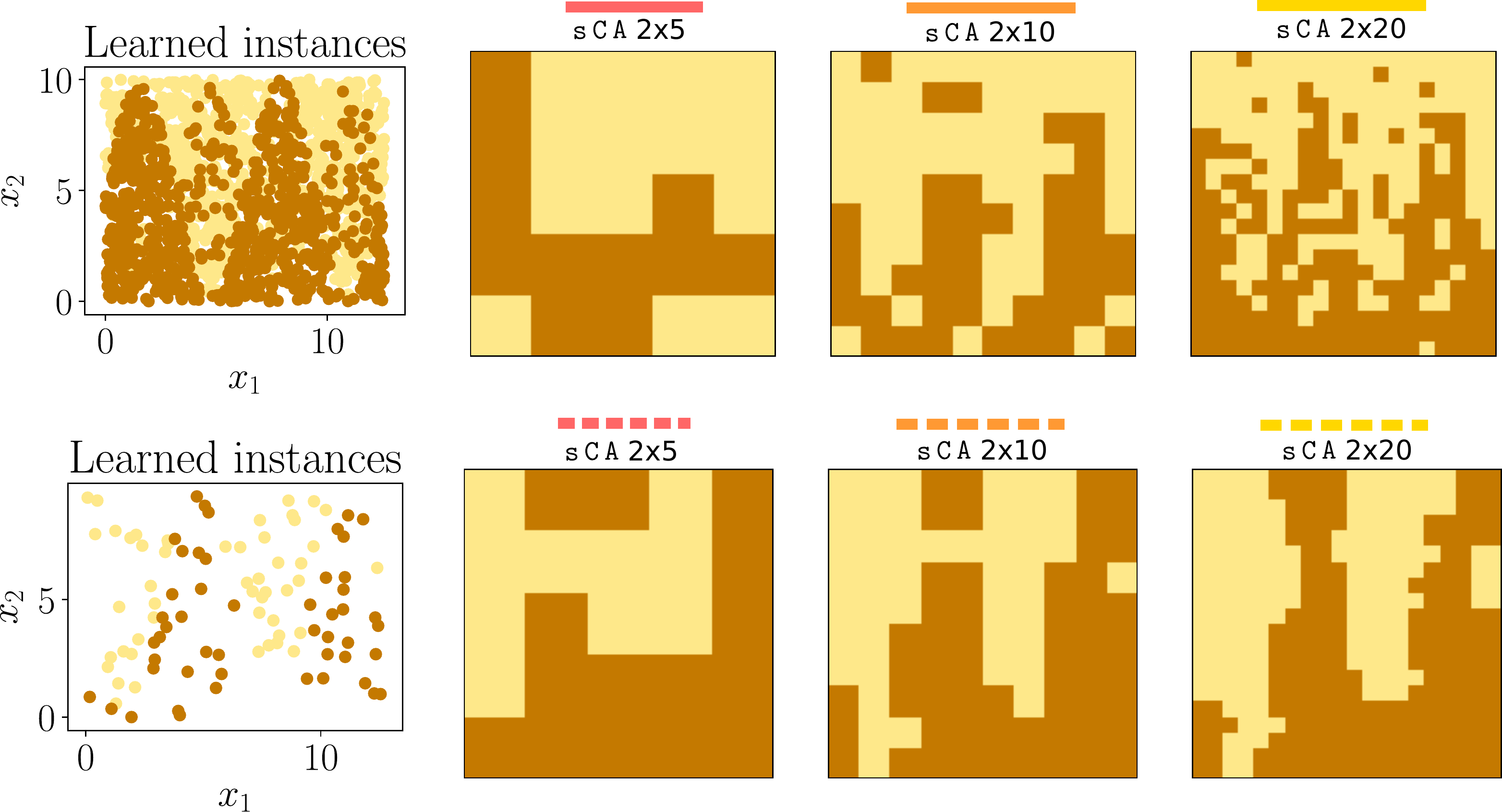} 
		\caption{}
		\label{fig_sineH33_CAs_3}
	\end{subfigure}
	\begin{subfigure}{0.5\textwidth}
		\centering
		% include first image		
		\includegraphics[width=\linewidth]{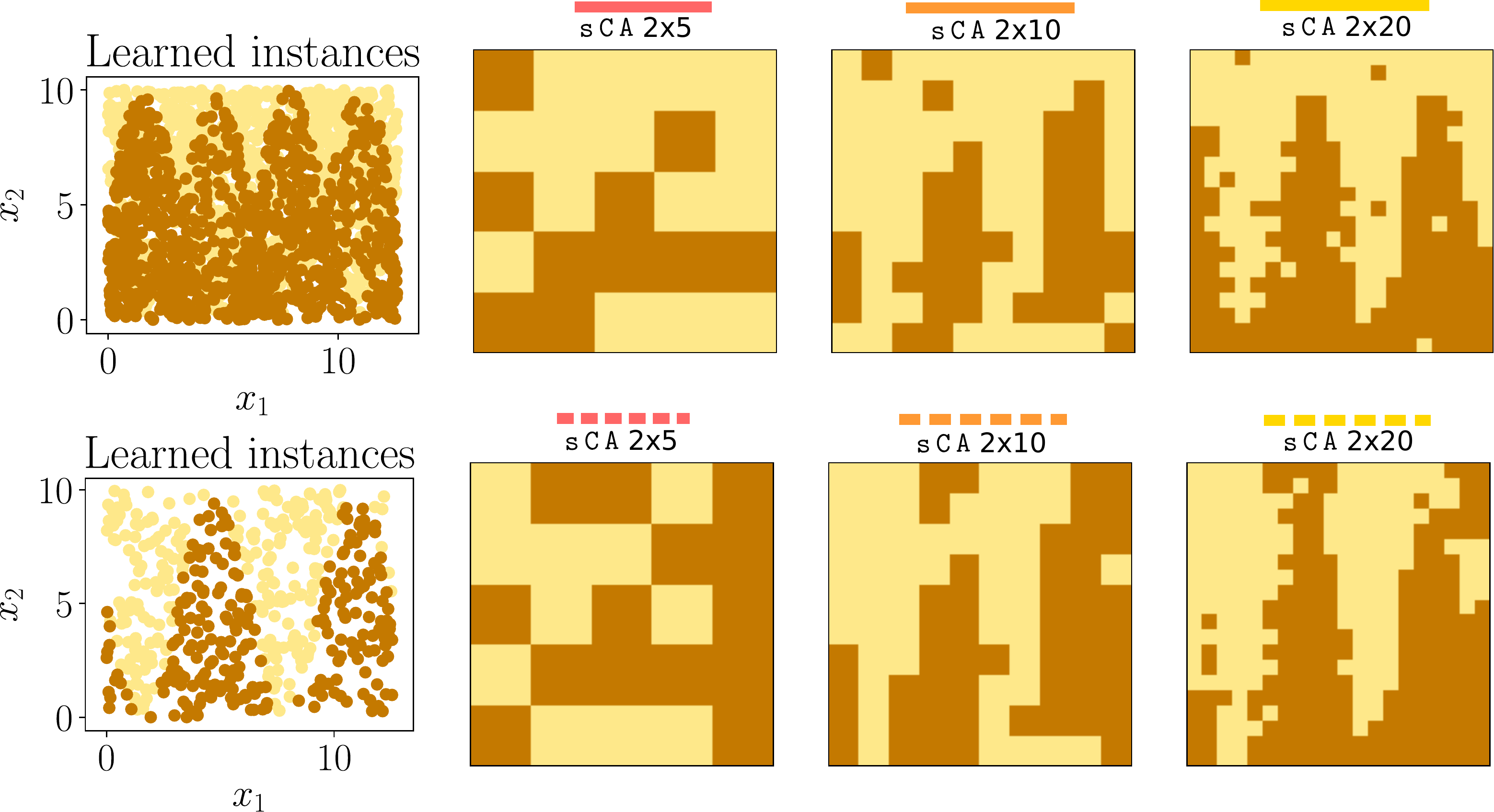} 
		\caption{}
		\label{fig_sineH33_CAs_4}
	\end{subfigure}
	\caption{Learning and adaptation of von Neumann's \texttt{sCA} of different grid sizes for the \texttt{SINEH} dataset, which exhibits a gradual drift at $t=1000$. The drift is detected at $t=1600$ and then the adaptation mechanism, which uses a window of instances $W=100$, is triggered. The interpretation of the remaining plots is the same than those in Figure \ref{fig_synths_circle31}.}
	\label{fig_synths_sineH33}
\end{figure}

Definitely yes. If we combining the definition of incremental learning suggested by \citep*{giraud2000note,lange2003formal} with Kuncheva's and Bifet's desiderata for non-stationary learning \citep*{kuncheva2004classifier,MOA-Book-2018}, incremental learning can be understood as the capacity to process and learn from data in an incremental fashion, and to deal with data changes (drifts) that may occur in the environment. Therefore, the properties that guide the creation of incremental learning algorithms \citep*{ditzler2012incremental} are:
\begin{itemize}[leftmargin=*]
	\item\textit{Learning new knowledge}: this feature is essential for online learners that have to be deployed over non-stationary settings. To provide evidence that \texttt{SCA} incorporates this characteristic, Figures \ref{fig_synths_circle31} and \ref{fig_synths_sineH31} summarize the performance results obtained over two synthetic datasets (\texttt{CIRCLE} and \texttt{SINEH}) with an abrupt drift occurring at $t=1000$, and assuming to have been detected at $t=1025$. The adaptation mechanism resorts to a window of $W=25$ past data instances; upon the detection of the drift, the entire knowledge contained in the cellular automata (i.e. the distribution of cell states) is erased, and the automata is seeded with the instances falling in the last $W$-sized window. Although the abrupt drift obliges to learn quickly the new concept and forgetting the old one, we observe in these plots that \texttt{sCA} excels at this task, yielding a prequential accuracy over time that does not get apparently affected by the change of concept.
	\item \textit{Preserving previous knowledge}: related to the previous property, gradual drifts require maintaining part of the previous concept over time, so that the evolved model leverages the retained knowledge during its transition to the new concept. This is clearly shown in Figures \ref{fig_synths_circle33} and  \ref{fig_synths_sineH33}, which both focus on \texttt{CIRCLE} and \texttt{SINEH} respectively, showing a gradual drift at $t=1000$, and assuming a drift detection at $t=1600$. The adaptation mechanism resorts to a window of $W=100$ past data instances. In particular, the prequential accuracy results and subplots therein included show that \texttt{sCA} can preserve the old concept while learning the new one through the implementation of a forgetting strategy.
	\item \textit{One-pass (incremental) learning}: through preceding sections we have elaborated on the incremental learning strategy of \texttt{sCA} (lines 30 to 33 in Algorithm \ref{alg_stream_automaton} ), by which it is able to learn one instance at a time without requiring any access to historical data.
	\item \textit{Dealing with concept drift}: Finally, we have seen in the previous experiments that \texttt{sCA} is capable of dealing with evolving conditions by using a drift detection and by implementing an adaptation scheme. We will later revolve around these results.
\end{itemize}

\newpage

\newpage

\newpage

\newpage

\newpage
\subsection{\texttt{RQ2}: Can \texttt{sCA} successfully adapt to evolving conditions?}

In a nutshell, the answer is yes. However, the success rate depends on the type of drift and the size $G$ of the CA grid. 

This is empirically shown in Figures \ref{fig_synths_circle31} and \ref{fig_synths_sineH31} for abrupt drifts, where we have applied an adaptation mechanism to a \texttt{sCA} of different grid sizes over the \texttt{CIRCLE} and \texttt{SINEH} datasets, respectively. Once the drift appears at $t=1000$ (solid vertical line) and is detected (dashdotted vertical line), we observe that \texttt{sCA} with and without adaptation mechanism perform equal at points \smash{\circleddefaults{b}} and \smash{\circleddefaults{c}}. But when the adaptation mechanism is triggered we see at point \smash{\circleddefaults{d}} ($25$ instances after drift detection and adaptation) and \smash{\circleddefaults{e}} (the end of the data stream) that $\texttt{sCA}$ using grids of $d\times G = 2\times10$ and $d\times G = 2\times20$ obtain better prequential accuracies when incorporating an adaptation mechanism. In the case of \texttt{CIRCLE} dataset, the adaptive versions of \texttt{sCA} obtain an average prequential accuracy of $0.873$ ($2\times10$) and $0.900$ ($2\times20$) versus their non-adaptive counterparts, which respectively score accuracies of $0.866$ and $0.880$. The same situation holds with the \texttt{SINEH} dataset, where again adaptive \texttt{sCA}s obtain average prequential accuracies of $0.773$ ($2\times10$) and $0.833$ ($2\times20$) against their non-adaptive versions ($0.765$ and $0.808$, respectively). The adaptive versions are able to forget quickly the old concept because they are reinitialized with a few instances ($25$) of the new concept, as opposed to the non-adaptive versions which keep the old concept along time (see Figures \ref{fig_circleG31_CAs_3} and \ref{fig_sineH31_CAs_3} ). At point \smash{\circleddefaults{e}} (Figures \ref{fig_circleG31_CAs_4} and \ref{fig_sineH31_CAs_4} ), we can observe some differences in the learning of both versions. In the case of the small grid size $2\times 5$, neither adaptive nor non-adaptive versions are able to learn the concepts properly and establishing the boundaries between classes, yielding a poor classification performance. Therefore, when the change from the old to the new concept is abrupt, the fast forgetting of the old concept becomes primordial \citep*{gama2013evaluating}. For a complete comparison with more datasets we refer to Table \ref{tab_panelA}. 

Now, we focus on gradual drifts of \texttt{CIRCLE} and \texttt{SINEH} dataset (Figures \ref{fig_synths_circle33} and \ref{fig_synths_sineH33} respectively), where a gradual drift also occurs at ($t=1,000$). Figures \ref{fig_circleG33_graph} and \ref{fig_sineH33_graph}, and Table \ref{tab_panelA}, show how both \texttt{sCA} with and without adaptation mechanism obtain almost the same mean prequential accuracy. As the old concept disappears slowly while the new one also does it slowly, the adaptive version with a window of instances ($w=100$) and the non-adaptive version do preserve the old concept while learn the new one. Even so, we can observe slight improvements of adaptive versions over the non-adaptive ones at points \smash{\circleddefaults{d}} and \smash{\circleddefaults{e}}. These differences are bigger in the case of $\texttt{sCA}$ with a $2\times20$ grid: $0.894$ for the adaptive version against $0.888$ for the non-adaptive version (\texttt{CIRCLE} at point \smash{\circleddefaults{d}}), $0.897$ for the adaptive version against $0.881$ for the non-adaptive version (\texttt{CIRCLE} at point \smash{\circleddefaults{e}}), $0.812$ for the adaptive version against $0.811$ for the non-adaptive version (\texttt{SINEH} at point \smash{\circleddefaults{d}}), and $0.820$ for the adaptive version against $0.811$ for the non-adaptive version (\texttt{SINEH} at point \smash{\circleddefaults{e}}). Again, in the case of a $2\times5$ grid, due to the small grid size neither adaptive nor non-adaptive versions are able to learn the concepts properly and establishing the boundaries between classes, achieving a poor classification performance. For the case of $\texttt{sCA} 2\times10$, because the drift is gradual, the grid size does not allow for capturing well enough the differences of both concepts. Finally, when the change from the old concept to the new one is gradual, an slow forgetting of the old concept becomes primordial \citep*{gama2013evaluating}.
\begin{table}[h!]
	\centering
	\resizebox{\textwidth}{!}{%
		\begin{tabular}{ccccccc}
			\toprule
			\multirowcell{2}{\textbf{Synthetic}\\\textbf{datasets}} & \multirow{2}{*}{\textbf{\texttt{sCA}}} & \multirow{2}{*}{$\mathbf{d\times G}$} & \multicolumn{4}{c}{$\mathbf{preACC(t)}$} \\
			\cmidrule(l){4-7}
			&  &  & \smash{\circleddefaults{c}} & \smash{\circleddefaults{d}} & \smash{\circleddefaults{e}} & \textbf{Average} \\ \midrule
			\multirow{6}{*}{\shortstack[c]{\texttt{CIRCLE}\\(abrupt)}} & \multirow{3}{*}{Adaptive} & $2\times5$ & $0.827$ & $0.822$ & $0.827$ & $0.794$ \\
			&  & $2\times10$ & $0.928$ & $0.918$ & $0.905$ & $0.873$ \\
			&  & $2\times20$ & $0.958$ & $0.943$ & $0.931$ & $0.900$ \\ \cmidrule(l){2-7}
			& \multirow{3}{*}{Non-adaptive} & $2\times5$ & $0.827$ & $0.822$ & $0.828$ & $0.795$ \\
			&  & $2\times10$ & $0.928$ & $0.915$ & $0.892$ & $0.866$ \\
			&  & $2\times20$ & $0.958$ & $0.939$ & $0.881$ & $0.880$ \\ \cmidrule(l){1-7}
			\multirow{6}{*}{\shortstack[c]{\texttt{CIRCLE}\\(gradual)}} & \multirow{3}{*}{Adaptive} & $2\times5$ & $0.827$ & $0.816$ & $0.820$ & $0.789$ \\
			&  & $2\times10$ & $0.928$ & $0.885$ & $0.890$ & $0.867$ \\
			&  & $2\times20$ & $0.958$ & $0.894$ & $0.897$ & $0.890$ \\ \cmidrule(l){2-7}
			& \multirow{3}{*}{Non-adaptive} & $2\times5$ & $0.827$ & $0.817$ & $0.820$ & $0.789$ \\
			&  & $2\times10$ & $0.928$ & $0.886$ & $0.891$ & $0.868$ \\
			&  & $2\times20$ & $0.958$ & $0.888$ & $0.881$ & $0.888$ \\ \cmidrule(l){1-7}
			\multirow{6}{*}{\shortstack[c]{\texttt{LINE}\\(abrupt)}} & \multirow{3}{*}{Adaptive} & $2\times5$ & $0.795$ & $0.784$ & $0.783$ & $0.754$ \\
			&  & $2\times10$ & $0.928$ & $0.912$ & $0.913$ & $0.868$ \\
			&  & $2\times20$ & $0.972$ & $0.959$ & $0.957$ & $0.911$ \\ \cmidrule(l){2-7}
			& \multirow{3}{*}{Non-adaptive} & $2\times5$ & $0.795$ & $0.785$ & $0.783$ & $0.754$ \\
			&  & $2\times10$ & $0.928$ & $0.905$ & $0.900$ & $0.861$ \\
			&  & $2\times20$ & $0.972$ & $0.953$ & $0.887$ & $0.883$ \\ \cmidrule(l){1-7}
			\multirow{6}{*}{\shortstack[c]{\texttt{LINE}\\(gradual)}} & \multirow{3}{*}{Adaptive} & $2\times5$ & $0.795$ & $0.721$ & $0.729$ & $0.731$ \\
			&  & $2\times10$ & $0.928$ & $0.870$ & $0.880$ & $0.855$ \\
			&  & $2\times20$ & $0.972$ & $0.910$ & $0.918$ & $0.898$ \\ \cmidrule(l){2-7}
			& \multirow{3}{*}{Non-adaptive} & $2\times5$ & $0.795$ & $0.722$ & $0.728$ & $0.731$ \\
			&  & $2\times10$ & $0.928$ & $0.871$ & $0.879$ & $0.855$ \\
			&  & $2\times20$ & $0.972$ & $0.903$ & $0.893$ & $0.895$ \\ 
			\bottomrule
		\end{tabular}%
		\hspace{1cm}
		\begin{tabular}{ccccccc}
			\toprule
			\multirowcell{2}{\textbf{Synthetic}\\\textbf{datasets}} & \multirow{2}{*}{\textbf{\texttt{sCA}}} & \multirow{2}{*}{$\mathbf{d\times G}$} & \multicolumn{4}{c}{$\mathbf{preACC(t)}$} \\
			\cmidrule(l){4-7}
			&  &  & \smash{\circleddefaults{c}} & \smash{\circleddefaults{d}} & \smash{\circleddefaults{e}} & \textbf{Average} \\ \midrule
			\multirow{6}{*}{\shortstack[c]{\texttt{SINEV}\\(abrupt)}} & \multirow{3}{*}{Adaptive} & $2\times5$ & $0.833$ & $0.822$ & $0.811$ & $0.777$ \\
			&  & $2\times10$ & $0.933$ & $0.920$ & $0.921$ & $0.877$ \\
			&  & $2\times20$ & $0.960$ & $0.947$ & $0.948$ & $0.907$ \\ \cmidrule(l){2-7}
			& \multirow{3}{*}{Non-adaptive} & $2\times5$ & $0.833$ & $0.823$ & $0.811$ & $0.777$ \\
			&  & $2\times10$ & $0.933$ & $0.912$ & $0.905$ & $0.867$ \\
			&  & $2\times20$ & $0.960$ & $0.939$ & $0.876$ & $0.879$ \\ \cmidrule(l){1-7}
			\multirow{6}{*}{\shortstack[c]{\texttt{SINEV}\\(gradual)}} & \multirow{3}{*}{Adaptive} & $2\times5$ & $0.833$ & $0.753$ & $0.756$ & $0.757$ \\
			&  & $2\times10$ & $0.933$ & $0.881$ & $0.891$ & $0.866$ \\
			&  & $2\times20$ & $0.960$ & $0.893$ & $0.906$ & $0.893$  \\ \cmidrule(l){2-7}
			& \multirow{3}{*}{Non-adaptive} & $2\times5$ & $0.833$ & $0.754$ & $0.757$ & $0.757$ \\
			&  & $2\times10$ & $0.933$ & $0.882$ & $0.892$ & $0.867$ \\
			&  & $2\times20$ & $0.960$ & $0.884$ & $0.879$ & $0.890$ \\ \cmidrule(l){1-7}
			\multirow{6}{*}{\shortstack[c]{\texttt{SINEH}\\(abrupt)}} & \multirow{3}{*}{Adaptive} & $2\times5$ & $0.514$ & $0.514$ & $0.501$ & $0.473$ \\
			&  & $2\times10$ & $0.822$ & $0.811$ & $0.806$ & $0.773$ \\
			&  & $2\times20$ & $0.889$ & $0.874$ & $0.877$ & $0.833$ \\ \cmidrule(l){2-7}
			& \multirow{3}{*}{Non-adaptive} & $2\times5$ & $0.514$ & $0.514$ & $0.503$ & $0.474$ \\
			&  & $2\times10$ & $0.822$ & $0.804$ & $0.790$ & $0.765$ \\
			&  & $2\times20$ & $0.889$ & $0.862$ & $0.822$ & $0.808$ \\ \cmidrule(l){1-7}
			\multirow{6}{*}{\shortstack[c]{\texttt{SINEH}\\(gradual)}} & \multirow{3}{*}{Adaptive} & $2\times5$ & $0.514$ & $0.513$ & $0.510$ & $0.478$ \\
			&  & $2\times10$ & $0.822$ & $0.756$ & $0.770$ & $0.756$ \\
			&  & $2\times20$ & $0.889$ & $0.812$ & $0.820$ & $0.818$ \\ \cmidrule(l){2-7}
			& \multirow{3}{*}{Non-adaptive} & $2\times5$ & $0.514$ & $0.514$ & $0.511$ & $0.478$ \\
			&  & $2\times10$ & $0.822$ & $0.763$ & $0.774$ & $0.757$ \\
			&  & $2\times20$ & $0.889$ & $0.811$ & $0.811$ & $0.817$ \\ \bottomrule			
		\end{tabular}%
	}
	\caption{Comparative results of \texttt{sCA} with and without adaptive mechanism with different grid sizes for synthetic \texttt{CIRCLE}, \texttt{LINE}, \texttt{SINEV} and \texttt{SINEH} datasets. The prequential accuracy is measured at points \smash{\circleddefaults{c}} (before the drift occurs), \smash{\circleddefaults{d}} (after drift occurs and adaptation mechanism has been triggered), and \smash{\circleddefaults{e}} (final). The last column amounts to the mean prequential accuracy averaged over the duration of the whole stream.}
	\label{tab_panelA}
\end{table}

In light of the above claim, we can finally confirm that in case of abrupt drifts we should opt for small grid sizes $G$, which is better in terms of computational cost. However, for gradual drifts the grid should be made finer in order to emphasize the differences between adaptive and non-adaptive versions. For the sake of space, only the cases of \texttt{CIRCLE} and \texttt{SINEH} datasets are presented in Figures \ref{fig_synths_circle31} (abrupt drift) and \ref{fig_synths_circle33} (gradual drift), and \ref{fig_synths_sineH31} (abrupt drift) and \ref{fig_synths_sineH33} (gradual drift), respectively. The results for all the synthetic datasets can be found in Table \ref{tab_panelA}.

\subsection{\texttt{RQ3}: Is \texttt{\algname} competitive in comparison with other consolidated OLMs of the literature?}

\begin{figure}[h!]
	\begin{subfigure}{\textwidth}
		\centering
		% include first image		
		\includegraphics[width=0.77\linewidth]{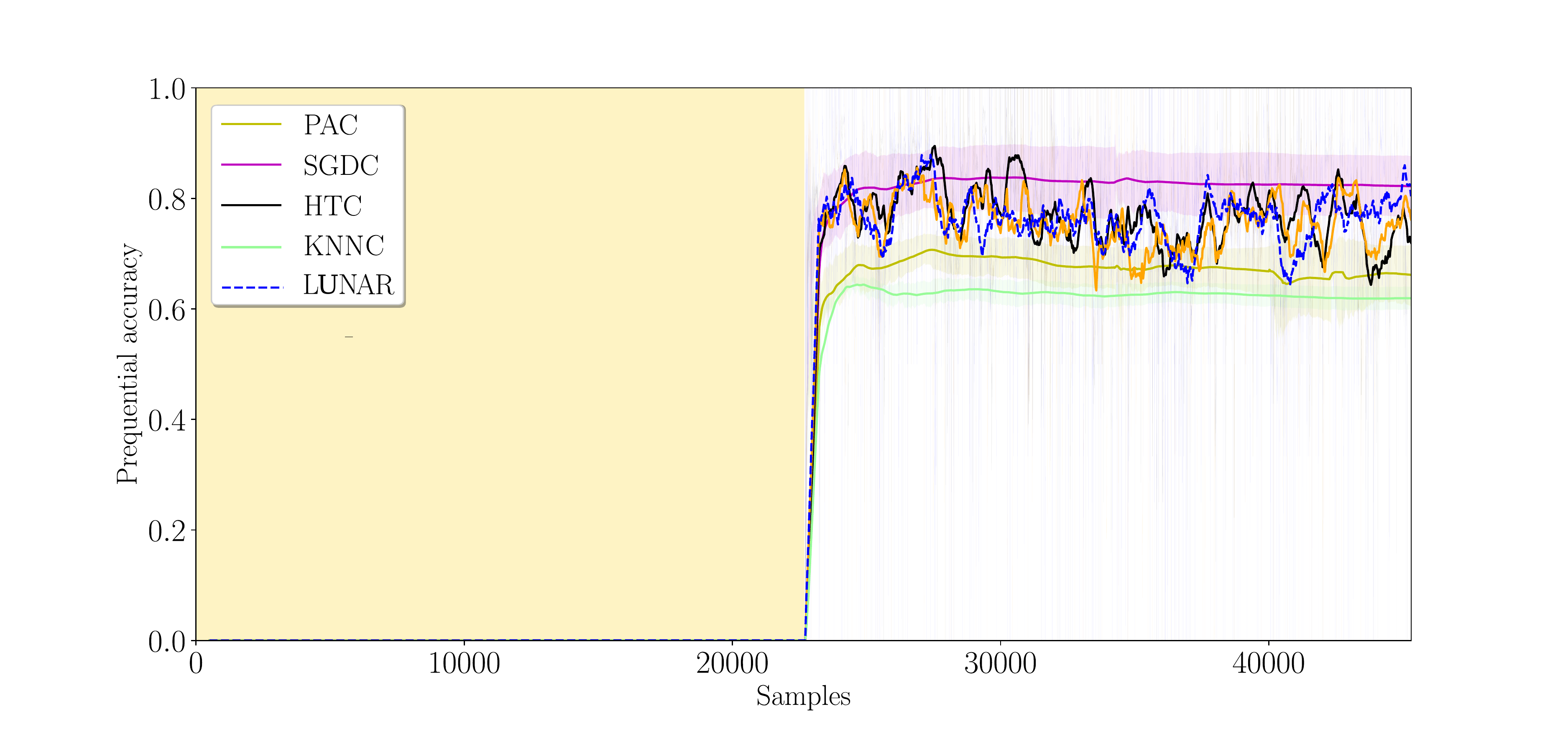} 
		\caption{\texttt{ELEC2} dataset}
		\label{fig_real_elec2}
	\end{subfigure}
	\begin{subfigure}{\textwidth}
		\centering
		% include first image		
		\includegraphics[width=0.77\linewidth]{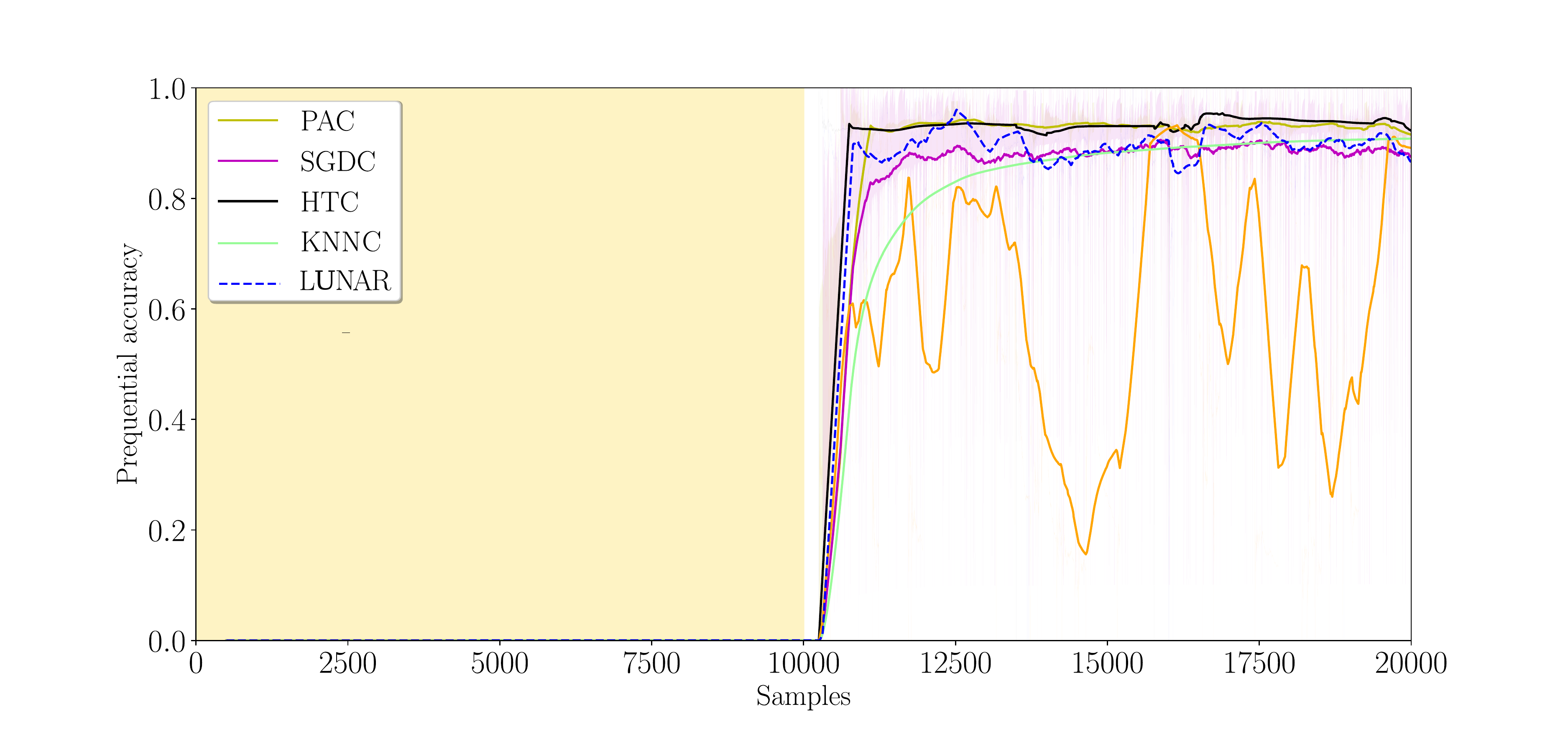} 
		\caption{\texttt{GMSC} dataset}
		\label{fig_real_gmsc}
	\end{subfigure}
	\begin{subfigure}{\textwidth}
		\centering
		% include first image		
		\includegraphics[width=0.77\linewidth]{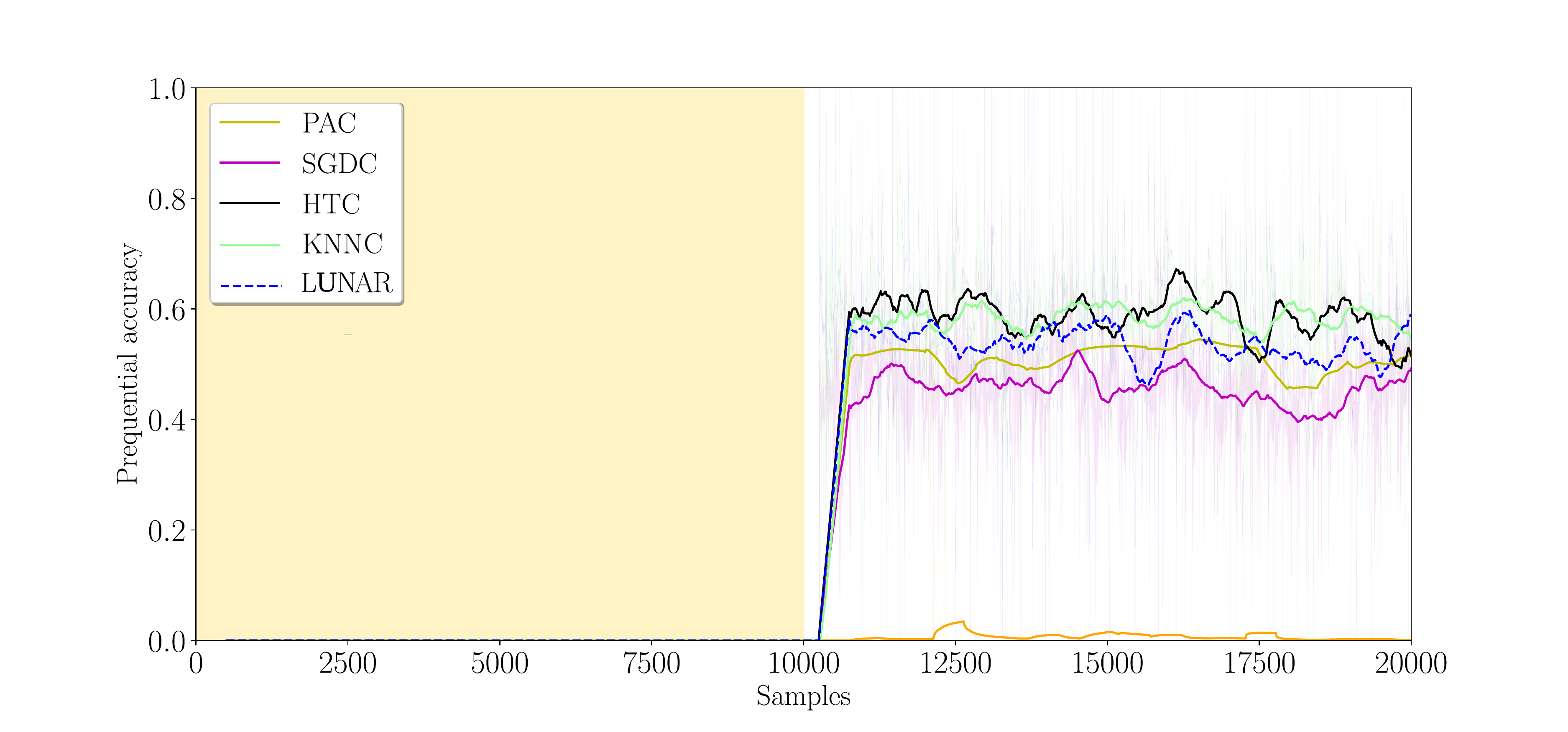} 
		\caption{\texttt{POKER-HAND} dataset}
		\label{fig_real_poker}
	\end{subfigure}
	\caption{Prequential accuracy $preACC(t)$ for \texttt{\algname} and the OLMs under consideration in the selected real-world datasets. The first $50\%$ of the dataset is for preparatory instances, while the rest is adopted as streaming \textit{test-then-train} instances. A moving window of $500$ instances has been applied in order to smooth out short-term fluctuations and to furnish a more friendly visualization.}
	\label{fig_reals}
\end{figure}

In Table \ref{real_results} and Figure \ref{fig_reals} we analyze the competitiveness of \texttt{\algname} in terms of classification performance for the considered real-world datasets. In the \texttt{ELEC2} dataset, only \texttt{SGDC} ($0.820$) outperforms \texttt{\algname} ($0.763$), and \texttt{HTC} shows almost the same score ($0.766$). In case of the \texttt{GMSC} dataset, only \texttt{HTC} ($0.910$) and \texttt{PAC} ($0.891$) outperforms \texttt{\algname} ($0.869$). For the \texttt{POKER-HAND} dataset, only \texttt{HTC} ($0.573$) and \texttt{KNNC} ($0.567$) outperforms \texttt{\algname} ($0.527$). 

Following the evaluation method used in \citep*{bifet2010fast}, the column \textit{Global mean $\mathbf{preACC}$} of Table \ref{real_results} averages the results in all datasets, and shows how \texttt{\algname} ($0.719$) is the second best method, only surpassed by \texttt{HTC} ($0.749$), which is arguably one of the best stream learners in the field. After the analysis of the results, we can confirm that effectively \texttt{\algname} has turned into a very competitive stream learning method.
\begin{table}[h!]
	\centering
	\resizebox{0.8\textwidth}{!}{%
		\begin{tabular}{@{}cccc@{}}
			\toprule
			\textbf{Method} & \textbf{Dataset} & $\mathbf{preACC}$ (mean$\pm$std) & \textbf{Global mean $\mathbf{preACC}$} \\ \midrule
			\multirow{3}{*}{\texttt{\algname}} & \texttt{ELEC2} & $0.763\pm0.0$ & \multirow{3}{*}{$\textbf{0.719}\pm\textbf{0.0}$} \\
			& \texttt{GMSC} & $0.869\pm0.0$ &  \\
			& \texttt{POKER-HAND} & $0.527\pm0.0$ &  \\ \cmidrule(l){1-4}
			\multirow{3}{*}{\texttt{SGDC}} & \texttt{ELEC2} & $0.820\pm0.058$ & \multirow{3}{*}{$0.703\pm0.068$} \\
			& \texttt{GMSC} & $0.845\pm0.085$ &  \\
			& \texttt{POKER-HAND} & $0.446\pm0.061$ &  \\ \cmidrule(l){1-4}
			\multirow{3}{*}{\texttt{HTC}} & \texttt{ELEC2} & $0.766\pm0.0$ & \multirow{3}{*}{$\textbf{0.749}\pm\textbf{0.0}$} \\
			& \texttt{GMSC} & $0.910\pm0.0$ & \\
			& \texttt{POKER-HAND} & $0.573\pm0.0$ & \\ \cmidrule(l){1-4}
			\multirow{3}{*}{\texttt{PAC}} & \texttt{ELEC2} & $0.670\pm0.043$ & \multirow{3}{*}{$0.685\pm0.030$} \\
			& \texttt{GMSC} & $0.891\pm0.032$ &  \\
			& \texttt{POKER-HAND} & $0.496\pm0.015$ &  \\ \cmidrule(l){1-4}
			\multirow{3}{*}{\texttt{KNNC}} & \texttt{ELEC2} & $0.621\pm0.019$ & \multirow{3}{*}{$0.670\pm0.029$} \\
			& \texttt{GMSC} & $0.822\pm0.014$ &  \\
			& \texttt{POKER-HAND} & $0.567\pm0.055$ &  \\ \bottomrule
		\end{tabular}%
	}
	\caption{Comparative results of \texttt{\algname} with other stream learners in real-world datasets. The column $\mathbf{preACC}$ (mean$\pm$std) denotes the mean prequential accuracy of every method in each dataset. The column \textbf{Global mean $\mathbf{preACC}$} presents the mean prequential accuracy for all datasets of each method, and serves as a reference for a global comparison. Global results of the two best methods are marked in bold.}
	\label{real_results}
\end{table}

\subsection{Final Observations, Remarks and Recommendations}

Once the three research questions have been thoroughly answered, it is worth discussing some general recommendations related to \texttt{\algname}. When designing a \texttt{sCA} for non-stationary scenarios, we should consider that generally we will assign one grid dimension to each feature of the dataset, thus as many dimensions in the grid and as many cells per feature, as much computational cost and more processing time will be required. Specifically, provided that a problem with $d$ dimensions is under target, and given a granularity (size) of the grid given by $G$, the worst-case complexity of predicting the class of a given test instance $\mathbf{X}_t$ is given by $\mathcal{O}(G^d)$, which is the time taken by a single processing thread to explore all cells of the $d$-dimensional grid of cells and discriminate the cell enclosing $\mathbf{X}_t$. Due to the exponential complexity, we recommend the use of \texttt{sCA} (and CA in pattern recognition in general) in datasets with a low number of features. Nevertheless, the search process over the grid's cells can be easily parallelized, hence allowing for fast prediction and cell updating speeds.

We also pause at the \textit{streamified} version of the \texttt{KNNC} model, whose similarity to \texttt{\algname} calls for a brief comparison among them. In essence, both learning models rely on the concept of neighborhood in the feature space, induced by either the selected measure of similarity among instances (\texttt{KNNC}) or the arrangement of cells in a grid (\texttt{\algname}). We have seen in Table \ref{real_results} that our proposed method overcomes the results of \texttt{KNNC} in \texttt{ELEC2} and \texttt{GMSC} datasets. Here it is worth mentioning that in order to have a fair comparison between both methods, the number of nearest neighbors to search for in the \texttt{KNNC} method has been properly optimized before the streaming process runs. Furthermore, the maximum size of the window storing the last viewed instances of \texttt{KNNC} has been set equal to the $W$ parameter in \texttt{\algname}. The capability of the \texttt{sCA} grid's cells to reflect and maintain the prevailing knowledge in the stream renders a superior performance than that offered by the \texttt{KNNC} method when computing the similarity over the $W$-sized window of past instances. 

\section{Conclusions and Future Work}\label{concs}

Cellular automata have been successfully applied to different real-world applications since their inception several decades ago. Among them, pattern recognition has been proven to be a task in which the self-organization and modeling capabilities of cellular automata can yield significant performance gains. This work has built upon this background to explore whether such benefits can also be extrapolated to real-time machine learning under non-stationary conditions, which is arguably among the hottest topics in Data Science nowadays. Under the premises of real-time settings, aspects such as complexity and adaptability to changing tasks are also important modeling design drivers that add to the performance of the model itself. 

The algorithm introduced in this work provides a new perspective in the stream learning scene. We have proposed a cellular automaton able to learn incrementally and capable of adapting to evolving environments (\texttt{\algname}), showing a competitive classification performance when is compared with other reputed state-of-the-art algorithms. \texttt{\algname} contributes to the discussion on ways to use cellular automata for paradigms in which the computation effort relies on a network of simple, interconnected devices with very low processing capabilities and constrained battery capacity (e.g. Utility Fog, Smart Dust, MEMS or Swarms). Under these circumstances, learning algorithms should be embarked in miniaturized devices running on low-power hardware with limited storage.

\texttt{\algname} has shown a well performance in practice over the datasets considered in this study, with empirical evidences of its adaptability when mining non-stationary data streams. Future work will be devoted towards experimenting with other local rules or neighborhoods to determine their effects on these identified properties of cellular automata. Some preliminary experiments carried out off-line suggest that a \texttt{sCA} could also \textit{detect} drifts, which, along with their simplicity, paves the way towards adopting them in active strategies for stream learning in non-stationary setups. We have also conceived the possibility of configuring ensembles of \texttt{sCA}, wherein diversity among the constituent automata can be induced by very diverse means (e.g. online bagging, boosting or probabilistic class switching). We strongly believe that moving in this algorithmic direction may open up the chance of designing more powerful cellular automata with complementary, potentially better capabilities to deal with stream learning scenarios.

\section*{Acknowledgements} \label{acknow}

This work has received funding support from the ECSEL Joint Undertaking (JU) under grant agreement No 783163 (\textit{iDev40} project). The JU receives support from the European Union's Horizon 2020 research and innovation programme, national grants from Austria, Belgium, Germany, Italy, Spain and Romania, as well as the European Structural and Investment Funds. It has been also supported by the ELKARTEK program of the Basque Government (Spain) through the \textit{VIRTUAL} (ref. KK-2018/00096) research grant. Finally, Javier Del Ser has received funding support from the Consolidated Research Group \textit{MATHMODE} (IT1294-19), granted by the Department of Education of the Basque Government.

%\section*{Code availability}\label{code}

%\noindent The source code can be found at the following link:
%\newline
%\url{https://github.com/TxusLopez/Streaming_cellular_automata}.

\section*{References}\label{references}
\bibliographystyle{model5-names}
\bibliography{biblio}

\end{document}